# An invariance constrained deep learning network for PDE discovery

Chao Chen(陈超) [1,2,3], Hui Li(李惠) [1,2,3], Xiaowei Jin(金晓威) [1,2,3]

[1] *Key Lab of Smart Prevention and Mitigation of Civil Engineering Disasters of the Ministry of Industry and Information Technology, Harbin Institute of Technology, Harbin, 150090, China*
[2] *Key Lab of Structures Dynamics Behavior and Control of the Ministry of Education, Harbin Institute of Technology, Harbin, 150090, China*
[3] *Guangdong-Hong Kong-Macao Joint Laboratory for Data-Driven Fluid Mechanics and Engineering Applications, Harbin Institute of Technology (Shenzhen), Shenzhen, 518055, China*

**ABSTRACT**: The discovery of partial differential equations (PDEs) from datasets has attracted increased attention. However, the discovery of governing equations from sparse data with high noise is still very challenging due to the difficulty of derivatives computation and the disturbance of noise. Moreover, the selection principles for the candidate library to meet physical laws need to be further studied. The invariance is one of the fundamental laws for governing equations. In this study, we propose an invariance constrained deep learning network (ICNet) for the discovery of PDEs. Considering that temporal and spatial translation invariance (Galilean invariance) is a fundamental property of physical laws, we filter the candidates that cannot meet the requirement of the Galilean transformations. Subsequently, we embedded the fixed and possible terms into the loss function of neural network, significantly countering the effect of sparse data with high noise. Then, by filtering out redundant terms without fixing learnable parameters during the training process, the governing equations discovered by the ICNet method can effectively approximate the real governing equations. We select the 2D Burgers equation, the equation of 2D channel flow over an obstacle, and the equation of 3D intracranial aneurysm as examples to verify the superiority of the ICNet for fluid mechanics. Furthermore, we extend similar invariance methods to the discovery of wave equation (Lorentz Invariance) and verify it through Single and Coupled Klein-Gordon equation. The results show that the ICNet method with physical constraints exhibits excellent performance in governing equations discovery from sparse and noisy data.

| Nomenclature | | | |
|---|---|---|---|
| PDEs | Partial differential equations | $p$ | Pressure |
| GI | Galilean Invariance | $\Delta$ | Laplace operator |
| LI | Lorentz Invariance | $\upsilon$ | Viscosity |
| IC-Net | Invariance constrained neural network | $Re$ | Reynolds number |
| $x, y, z$ | Spatial coordinate | $\mathbf{u}_t$ | First order derivative of $u$ with respect to time $t$ |
| $t$ | Time coordinate | $u_{tt}$ | Second order derivative of $u$ with respect to time $t$ |
| $\bar{x}, \bar{y}, \bar{z}, \bar{t}$ | The moving coordinate | $u_x$ | Derivative of $u$ with respect to spatial coordinate $x$ |
| $\Omega$ | Computational domain | $B, R$ | The Lorentz boost matrix |
| $\mathbf{c}$ | Velocity of coordinate transformation | $\Theta$ | Initial library |
| $c_0$ | Speed of light | $\Theta_G$ | Galilean invariance library |
| $E$ | The fixed coordinate system | $\Theta_L$ | Lorentz invariance library |
| $\bar{E}$ | The moving coordinate system | $\Lambda$ | Sparse vector of initial library |
| $k$ | The maximum order of monomials in library | $\Lambda_L$ | Sparse vector of Galilean invariance library |
| $\mathbf{u}$ | Vector solution | $\Lambda_G$ | Sparse vector of Lorentz invariance library |
| $\hat{\mathbf{u}}$ | The approximation solution by neural network | $\lambda_G$ | The coefficient of term $\mathbf{u} \cdot \nabla \mathbf{u}$ |
| $\phi$ | Scalar field | $\lambda_L$ | The coefficient of term $\nabla^2 u$ |

## I. INTRODUCTION

Partial differential equations (PDEs) govern the physical world in concise and beautiful representation. The derivation of these PDEs is grounded in a careful examination of the physical world and a profound understanding of the fundamental laws governing matter and energy in nature. In the last decade, data-driven PDE discovery methodology has attracted more attention.[1-3]

The essence of PDE discovery through data-driven methodology is model regression, meaning that all items in the regression model are pre-determined, and the coefficients of the items are obtained through machine learning algorithms. Data-driven methods for discovering PDEs can be categorized into three classes according to the determination strategy of candidate items in a library. The first category of methods is to construct an over-complete candidate library, and then sparsity is addressed through regularization of the loss function to fix overfitting problem. For instance, the Sparse Identification of Nonlinear Dynamics (SINDy) method, proposed by Brunton et al.[4-6] has demonstrated success in the parsimonious discovery of ordinary differential equations (ODEs), later extends to various dynamic systems, including the chaotic Lorentz system, parameterized dynamic systems and systems with external disturbance. Then, Rudy et al.[7] introduced the Sequential Threshold Ridge Regression (STRidge) method for PDE discovery from spatiotemporal datasets. Furthermore, STRidge is employed in the discovery of governing equations across different scales. Zhang and Ma[8] utilized direct simulation Monte Carlo (DSMC) to simulate the microscopic molecular movements and discovered the underlying macroscopic governing equations from the simulation datasets. For the discovery of PDEs with spatially or temporally varying coefficients, Rudy et al.[9] discovered active terms from candidates and obtained the time or space dependence of coefficients through group sparsity. Due to $L^1$ regularization with better sparse optimization capability, Schaefer[10] validated the effectiveness of the $L^1$ regularized least-squares method for learning active terms in PDEs. Chang and Zhang[11] further employed $L^1$ regularization to unveil the groundwater flow equation and the contaminant transport equation from the datasets. Prior to the construction of the library, the computation of derivatives is a prerequisite to derive fundamental candidate terms. However, traditional numerical differentiation methods may result in relatively large errors, particularly for very sparse and noisy data. To reduce these errors, Goyal and Benner[12] combined the numerical integration framework with candidates library to discover ODEs without the requirement of derivatives. This approach has been proven to fix sparse and noisy data. Xu et al.[13] used the automatic differentiation technique embedded in neural networks to calculate derivatives before construction of the library for PDE discovery. Berg and Nyström[14] further affirmed the superiority of the automatic differentiation technique to process complex datasets. Additionally, Both et al.[15] have directly incorporated the library into the physics-informed loss functions. To further obtain concise governing equations, Chen et al.[16,17] applied sparse regression to filter out unnecessary terms. These methods above need to build an over-complete library, which may not meet or even be against the physical laws of governing equations.

The second category of methods is to generate candidates through combination of basic terms without construction of an over-complete library. Xu et al.[18,19] utilized mutation and crossover in gene expression programming to generate active terms, facilitating the discovery of governing equations without the necessity of an overcomplete library. Based on the groundwork, Zeng et al.[20] applied gene expression programming to unveil the macroscopic governing equations of viscous gravity currents from microscopic simulation data. Xing et al.[21] also used gene expression programming to discover governing equations hidden in the complex fluid dynamics from molecular simulation datasets. To further optimize the coefficients of the PDEs, the active terms discovered through gene expression programming are incorporated into the physics-informed neural network to enhance the accuracy of results.[22] For PDEs with spatially or temporally varying coefficients, Xu et al.[23] first employed gene expression programming to identify terms contained in the PDEs, subsequently, they utilized a stepwise adjustment strategy to get the general form of the spatially or temporally varying coefficients. Besides, Symbolic networks possess the capability not only to generate candidates from basic terms but also to perform predictions. Long et al.[24] integrated symbolic networks with forward Euler temporal discretization to discover PDE models and facilitate the prediction of dynamical behavior for a relatively long time. Nevertheless, it contains numerous redundant terms, making it challenge to precisely identify the equation.. Simultaneously, the aforementioned gene expression programming can be used to discover PDEs in simple cases. However, with the increase in the number of variables, the search space of the genetic algorithm expands rapidly, thereby effiency is not good enough.[25,26]

In the third category of methods, candidates are fixed without any redundant terms. Raissi et al.[27-29] used physics-informed neural networks to learn the coefficients of active terms by directly embedding the PDEs into the loss function as constraints. In this scenario, all effective terms of dynamical systems are known, while the coefficients of certain terms remain unknown. To determine the effective terms, Cao and Zhang[30] integrated dimensional analysis and the direction of the equations to ascertain candidate terms for the dynamical system. This approach proved successful in the discovery of governing equations from the flow field data of the Karmen vortex street. However, this kind of methods needs to have an insight on the systems represented by PDEs to achive the fixed candidates without redundant terms, which is not straightforward.

Incorporation of physical laws can be a favorite to discover correct PDEs. We thus propose the invariance constrained deep learning networks (ICNet) to discover various PDEs from sparse and noisy data. Given that the invariance is a crucial property of physical laws, we first perform invariance transformation on the candidates and the general form of PDEs. The candidate library, satisfying the invariance, is derived and embedded into the loss function of the neural network. Next, the ICNet is integrated with STRidge to filter out the redundant terms and decrease the equation residuals during the training process. Initially, the ICNet method is employed to discover governing equations within fluid mechanics, encompassing the two-dimensional Burgers equation, the equation of two-dimensional channel flow over an obstacle, and the equation of three-dimensional intracranial aneurysm. Subsequently, we extend a comparable invariance derivation approach to the realm of relativity, applying it to discover the Single and Coupled Klein-Gordon equation. The results demonstrate that the ICNet method can discover the governing equations with high accuracy from noisy and sparse datasets compared with existing methods and achieve state-of-the-art performance.

The remainder of this paper is outlined as follows. In Section 2, we provide an exposition on the construction of the invariance library and elucidate the architecture of ICNet. Section 3 encompasses an evaluation of the ICNet method through five illustrative examples. These examples include three scenarios within fluid mechanics (the two-dimensional Burgers equation, the equation of two-dimensional channel flow over an obstacle, and the equation of three-dimensional intracranial aneurysm). Furthermore, we extend our method to address relativistic wave equations (Single and Coupled Klein-Gordon equation). A comparative analysis with existing methods is also presented in this section. The study is summarized and concluded in Section 4.

## II. METHOD

### A. ICNet

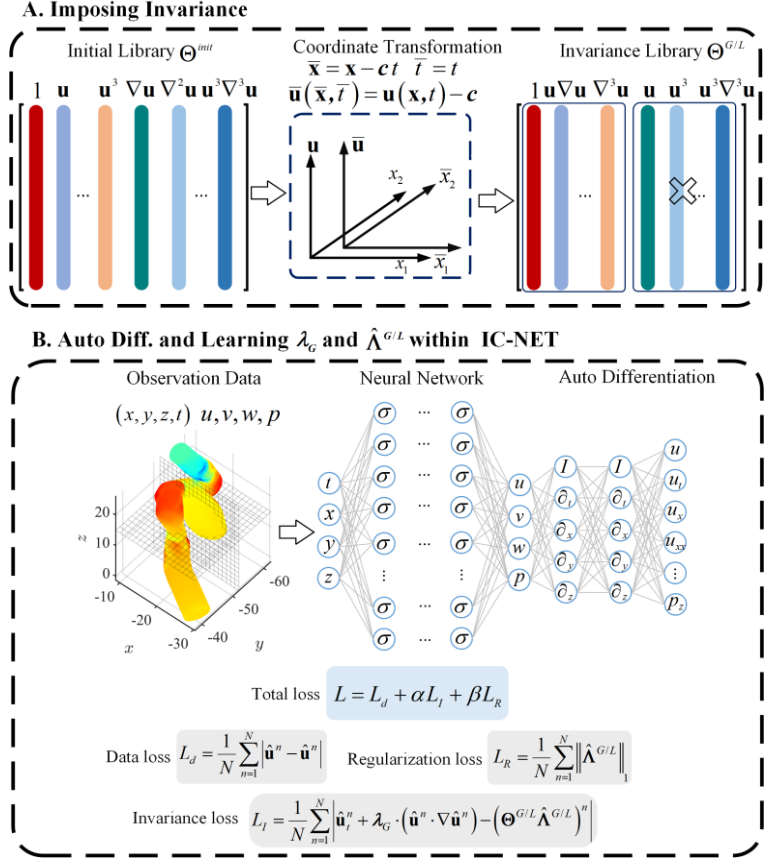

FIG. 1. Schematic of Invariance Constrained Deep Learning Network (ICNet)

The architecture of ICNet is shown in Fig. 1. ICNet is employed to approximate the observation data $\{\mathbf{u}(t, x, y, z), p(t, x, y, z)\}$, where the spatial and temporal coordinates serve as inputs, and the velocity and pressure field data are generated as outputs, i.e. $(t, x, y, z) \mapsto (u, v, w, p)$. The candidate terms are constructed based on the principles of time and space translation invariance (Galilean and Lorentz invariance). The incorporation of invariance into the loss function of ICNet is detailed as follows.

#### 1. Galilean invariance for PDE discovery

In Newtonian physics, Galilean invariance (GI) is a fundamental physical property. The governing equations are covariant, and their mathematical form is invariant with respect to (w.r.t.) Galilean transformation. The Galilean invariance should be naturally satisfied in discovery of PDE from a massive amount of data in the frame of Newtonian physics.

The generalized representation of partial differential equations in the Newtonian mechanism can be written as follows.

$$\mathbf{u}_t = N(\mathbf{u}, \nabla\mathbf{u}, \nabla^2\mathbf{u}, \mathbf{u}\cdot\nabla\mathbf{u}, \mathbf{u}^2\cdot\nabla\mathbf{u}, \dots; \xi) \tag{1}$$

where $\mathbf{u}(t,\cdot): \Omega \to \mathbb{R}^d$, $N(\mathbf{u}, \nabla\mathbf{u}, \nabla^2\mathbf{u}, \mathbf{u}\cdot\nabla\mathbf{u}, \mathbf{u}^2\cdot\nabla\mathbf{u}, \dots; \vartheta) \in \mathbb{R}^d$ is unknown and needs to be discovered from the given dataset, the $\xi$ denotes the parameters in $N(\cdot)$, the subscript $t$ denotes the partial derivation w.r.t. to time and $\nabla$ is the gradient operator w.r.t to spatial coordinate $\mathbf{x}$. Our goal is to discover the analytic form of governing equations from the given datasets $\mathbf{u}(\mathbf{x}, t)$ over a certain temporal and spatial domain, $\{\mathbf{u}(\mathbf{x},t): t \in \mathbb{R}, \mathbf{x} \in \Omega \subset \mathbb{R}^d\}$.

Based on the Galilean transformation $\bar{\mathbf{x}} = \mathbf{x} - \mathbf{c}t, \bar{t} = t$ and $\mathbf{c}$ is the relative velocity of two coordinates. Eq. (1) can be derived as,

$$\bar{\mathbf{u}}_{\bar{t}} - \mathbf{c} \cdot \nabla \bar{\mathbf{u}} = N(\bar{\mathbf{u}}, \nabla \bar{\mathbf{u}}, \nabla^2 \mathbf{u}, \bar{\mathbf{u}} \cdot \nabla \bar{\mathbf{u}}, \bar{\mathbf{u}}^2 \cdot \nabla \bar{\mathbf{u}}, \dots; \vartheta) \tag{2}$$

where $\bar{\mathbf{u}}$ is the variable after transformation.

Here we use $\Theta^k(\mathbf{u}, \nabla \mathbf{u}, \nabla^2 \mathbf{u}, \dots)$ to denote the overcomplete library with the degree of monomials no more than $k$ for discovery of equations. The principle of candidate terms is selected according to Galilean invariance

$$\overline{\Theta}^k(\bar{\mathbf{u}}, \nabla \bar{\mathbf{u}}, \nabla^2 \bar{\mathbf{u}}, \dots) = [1, \cdots, \bar{\mathbf{u}}^k, \nabla \bar{\mathbf{u}}, \cdots, \bar{\mathbf{u}}^{k-1} \cdot \nabla \bar{\mathbf{u}}, \nabla^2 \bar{\mathbf{u}}, \cdots] \tag{3}$$

where $\overline{\Theta}^k(\bar{\mathbf{u}}, \nabla \bar{\mathbf{u}}, \nabla^2 \bar{\mathbf{u}}, \dots)$ denotes the candidate terms. The key point to meet the Galilean invariance is that only partial derivative terms (e.g. $\nabla \mathbf{u}, \nabla^2 \mathbf{u}$) meet the requirement of Galilean invariance which can appear in the candidates, except for term $\mathbf{u} \cdot \nabla \mathbf{u}$ which counteract the term $\mathbf{c} \cdot \nabla \bar{\mathbf{u}}$ after Galilean transformation. While other rest terms containing $\mathbf{u}$ (e.g. $\mathbf{u}, \mathbf{u} \cdot \mathbf{u}, \mathbf{u} \times \mathbf{u}, \mathbf{u} \cdot \mathbf{u} \cdot \nabla \mathbf{u}$, etc.) cannot be included in the candidates because vector $\mathbf{u}$ would change with coordinate transformation, which is conflict with the Galilean invariance. The candidates meeting the invariance are rewritten as follows

$$\Theta_G^k(\mathbf{u}, \nabla \mathbf{u}, \nabla^2 \mathbf{u}, \dots) = [1, \mathbf{u} \cdot \nabla \mathbf{u}, \nabla \mathbf{u}, \nabla^2 \mathbf{u}, \nabla^3 \mathbf{u}, \cdots] \tag{4}$$

where $\Theta_G^k(\cdot)$ is the matrix of the candidates meeting the Galilean invariance.

Based on the discussion above, Eq. (1) can be reconstructed in the library as follows,

$$\mathbf{u}_t = \Theta_G^k \Lambda_G \tag{5}$$

where $\mathbf{u}_t$ denotes the derivative of multi-dimensional variable w.r.t time $t$, $\Lambda_G$ is sparse coefficient marix. Here we complete to embed Galilean invariance into the process of discovery of equations. It is needed to obtain the nonzero entries in the sparse matrix $\Lambda_G$.

## 2. Lorentz invariance for PDE discovery

We also apply similar invariance derivation results to relativistic wave equation. The variables considered here are scalars. In the context of relativity, Lorentz invariance (LI) is an important physical property. That is, governing equations are covariant and their mathematical form is invariant w.r.t. Lorentz transformation. For the PDE discovery from a massive amount of data under the frame of relativity, we should embed the Lorentz invariance into the discovery of PDE.

The general form of differential equations in relativity can be written as follows

$$u_{tt} = N(u, \nabla u, \nabla^2 u, u \nabla u, u^2 \nabla u, \dots; \xi) \tag{6}$$

where $u(t,\cdot): \Omega \to \mathbb{R}^d$, $N(u, \nabla u, \nabla^2 u, u \nabla u, u^2 \nabla u, \dots; \vartheta) \in \mathbb{R}^d$ is unknown and needs to be discovered from the given dataset, $\xi$ are the parameters in $N(\cdot)$, the subscript $t$ denote the partial derivation w.r.t time and $\nabla$ is the gradient operator w.r.t to spatial coordinate $\mathbf{x}$.

Based on the Lorentz transformation $\bar{\mathbf{x}} = \mathbf{x} + (\gamma - 1)\frac{\mathbf{c}(\mathbf{c} \cdot \mathbf{x})}{c^2} - \gamma \mathbf{c}t, \bar{t} = \gamma t - \frac{\gamma(\mathbf{c} \cdot \mathbf{x})}{c^2}$ and $\mathbf{c}$ is the relative velocity of two coordinates. Eq. (6) can be derived as,

$$\gamma^2\big(\bar{u}_{\bar{t}\bar{t}} - 2\mathbf{c} \cdot \nabla \bar{u}_{\bar{t}} + tr(\mathbf{c}\mathbf{c} \cdot [\nabla(\nabla \bar{u})])\big) = N\big(\bar{u}, R^1 \cdot \nabla \bar{u} - B_0^1 \bar{u}_{\bar{t}}, \dots, R^d \cdot \nabla \bar{u} - B_0^d \bar{u}_{\bar{t}} \dots; \vartheta\big) \tag{7}$$

where $\bar{u}$ is the variable after transformation, B is the Lorentz boost matrix, $R$ is the spatial component of Lorentz boost matrix, $\gamma$ is Lorentz factor and the $tr(\cdot)$ calculates the trace of a matrix.

Here we use $\Theta^k(u, \nabla u, \nabla^2 u, \dots)$ to denote the overcomplete library with the degree of monomials no more than $k$ for discovery of equations. The principle of candidate terms is selected according to Lorentz invariance.

$$\overline{\Theta}^k(\bar{u}, R^1 \cdot \nabla \bar{u} - B_0^1 \bar{u}_{\bar{t}}, \dots) = [1, \cdots, \bar{u}^k, R^1 \cdot \nabla \bar{u} - B_0^1 \bar{u}_{\bar{t}}, \dots, R^d \cdot \nabla \bar{u} - B_0^d \bar{u}_{\bar{t}}, \cdots] \tag{8}$$

where $\overline{\Theta}^k(\bar{u}, R^1 \cdot \nabla \bar{u} - B_0^1 \bar{u}_{\bar{t}}, \dots)$ denotes the candidate terms after the Lorentz transformation. The key point to meet the Lorentz invariance is that terms without partial derivatives (e.g. $u, u^2$) meet the requirement of Lorentz invariance which can appear in the candidates, except for term $\nabla^2 u$ which counteracts the terms generated by $u_{tt}$ after transformation. While other rest terms containing partial derivatives (e.g. $\nabla u, u\nabla u$) cannot be included in the candidates. The candidates meeting the Lorentz invariance are rewritten as follows

$$\Theta_L^k(u, \nabla u, \nabla^2 u, \ldots) = [1, \nabla^2 u, u, u^2, u^3, \cdots] \qquad (9)$$

where $\Theta_L^k(\cdot)$ is the matrix of the candidates meeting the Lorentz invariance.

Based on the findings above, Eq. (6) can be rewritten as follows

$$\mathbf{u}_{tt} = \Theta_L^k \Lambda_L \qquad (10)$$

where $\mathbf{u}_{tt}$ denotes the second-order partial derivatives w.r.t to time, $\Lambda_L$ is the sparse coefficient matrix. The $\Theta_L^k$ denotes the matrix of the library meeting the Lorentz invariance.

To this end, we have proposed the method to embed invariance into the discovery of equations by redefining the library instead of complicated and unprincipled candidates. Next, we would implement the requirement of invariance within neural network for discovery of equations.

### 3. Loss function with invariance

Considering that the powerful representation capabilities and automatic differentiation technique of neural network,[15,16] we use it to solve the model regression task in (1) and (6). We further enhance the interpretability of neural network learning by embedding invariance. The fully connected feed forward neural network is used to approximate the variable $\mathbf{u}$ of the datasets, which is also the output of the network. While the input of neural network is the spatial and temporal coordinate $(x, y, z, t)$. Then the automatic differentiation technique of network is used to compute derivatives. The derivatives and variable $\hat{\mathbf{u}}$ approximated by neural work are used to construct the invariance constrained candidates $\Theta_G^k$ and $\Theta_L^k$ which are used to form the physical loss function.

Therefore, the loss function of neural network is consisted of three components

$$\mathcal{L} = \mathcal{L}_d + \alpha \mathcal{L}_I + \beta \mathcal{L}_R \qquad (11)$$

where $\mathcal{L}_{data}$ is the data loss, and $\mathcal{L}_R$ is regularization loss of $\Lambda = \Lambda_G$ or $\Lambda_L$,

$$\mathcal{L}_d = \frac{1}{N} \sum_{n=1}^{N} \|\mathbf{u}^n - \hat{\mathbf{u}}^n\|_2^2 \qquad (12)$$

$$\mathcal{L}_R = \|\Lambda\|_1 \qquad (13)$$

where $\mathbf{u}$ is the given true datasets, $\hat{\mathbf{u}}$ is the corresponding value approximated by neural network and $N$ is the number of training data. $\alpha$ and $\beta$ are the weight coefficients.

The $\mathcal{L}_I$ is the physical loss meeting the requirement of invariance. According to the property of two invariance above, we propose two form of physical loss function $\mathcal{L}_{GI}$ and $\mathcal{L}_{LI}$, respectively,

$$\mathcal{L}_{GI} = \frac{1}{N} \sum_{n=1}^{N} \left\| \hat{\mathbf{u}}_t^{\ n} + \lambda_G \cdot (\hat{\mathbf{u}}^n \cdot \nabla \hat{\mathbf{u}}^n) - \left(\Theta_G^k \Lambda_G\right)^n \right\|_2^2 \qquad (14)$$

$$\mathcal{L}_{LI} = \frac{1}{N} \sum_{n=1}^{N} \left\| \hat{\mathbf{u}}_{tt}^{\ n} + \lambda_L \cdot (\nabla^2 \hat{\mathbf{u}}^n) - \left(\Theta_L^k \Lambda_L\right)^n \right\|_2^2 \qquad (15)$$

Because the $\mathbf{u} \cdot \nabla \mathbf{u}$ and $\nabla^2 u$ must exist according to the requirement of invariance, we take them out from the library $\Theta_G^k$ and $\Theta_L^k$, respectively. The coefficients $\lambda_G$ and $\lambda_L$ can be learned without being suppressed in the regularization loss and improve the accuracy of discovery during the training of neural network. Therefore, the trainable parameters contain neural network parameters and the coefficients of candidates $\lambda_{G/L}$ and $\Lambda_{G/L}$. The model regression task (1) and (6) can be implemented by minimizing the loss function $\mathcal{L}_{GI}$ or $\mathcal{L}_{LI}$.

## B. Training of ICNet

Thus, the problem of discovery of equations defined in Eq. (1) and Eq. (6) equals to solve optimization problem defined in Eq. (16).

$$\{\hat{\theta}^k, \hat{\lambda}^k, \widehat{\Lambda}^k\} = argmin_{\{\theta,\lambda,\Lambda\}}\{\mathcal{L}_d + \mathcal{L}_I + \mathcal{L}_R\} \tag{16}$$

In the context of a general gradient descent algorithm, the iterative optimization of ICNet parameters can be formulated as:

$$\{\theta^{k+1}, \lambda^{k+1}, \Lambda^{k+1}\} = \{\theta^k, \lambda^k, \Lambda^k\} - \eta\nabla_{\{\theta,\lambda,\Lambda\}}\mathcal{L}_d - \eta\alpha\nabla_{\{\theta,\lambda,\Lambda\}}\mathcal{L}_I - \eta\beta\nabla_{\{\theta,\lambda,\Lambda\}}\mathcal{L}_R \tag{17}$$

where $\theta$ denotes the trainable parameters of the neural network, $\lambda$ denotes $\lambda_G$ or $\lambda_L$, $\Lambda$ denotes $\Lambda_G$ or $\Lambda_L$, $k$ is the iteration step, and $\eta$ is the learning rate.

We train the ICNet by optimize the learnable parameters including network parameters $\theta$, the coefficients of candidates $\lambda_{G/L}$ and $\Lambda_{G/L}$. We initially employ the Adam optimizer for pretraining the neural network to obtain better initial values of the trainable parameters.[31] Subsequently, the L-BFGS-B optimizer is utilized to accelerate the discovery and enhance the accuracy of coefficients.[32] The learning rate is set to be $10^{-3}$ uniformly. If the data cannot be trained within one batch, we will consistently employ the Adam optimizer. $L^1$ regularization is implemented during the optimization process of the neural network to enforce the sparsity of coefficients. Consequently, we will acquire a portion of coefficients with small values, and this fraction corresponds to redundant candidates in the library. To filter out redundant terms in $\Theta_{G/L}^k$, we utilize Sequential Threshold Ridge Regression (STRidge)[7] every $K$ iteration steps during the training process, facilitating continuous optimization of the trainable parameters. Simultaneously, we adjust $\beta$ to gradually enhance sparsity. The key operation of STRidge for filtering out redundant terms is accomplished by setting adaptive threshold tolerance,[7] and the filtered element $\hat{\lambda}_{ij}^k$ in the $i$th row and $j$th column of $\widehat{\Lambda}^k$ is

$$\hat{\lambda}_{ij}^k = \begin{cases} 0, & if\ \lambda_{ij}^k < tol \\ \lambda_{ij}^k, & else \end{cases} \tag{18}$$

where $\lambda_{ij}^k$ is the element of the coefficient matrix obtained by training ICNet, $tol$ is the threshold tolerance, and more details can be found in the literature.[7] Coefficients smaller than the $tol$ are filtered, while coefficients larger than $tol$ are retained. The training process would be finished until the size of candidates $\Theta_{G/L}^k$ remains unchanged.

## III. RESULTS

In this section, we first validate the accuracy and robustness of ICNet in the discovery of governing equations in fluid mechanics. (the two-dimensional Burgers equation, the equation of two-dimensional channel flow over an obstacle, and the equation of three-dimensional intracranial aneurysm). Subsequently, ICNet is extended to the realm of relativity, and its accuracy and robustness are similarly verified (Single and Coupled Klein-Gordon equation).

### A. Numerical examples for fluid mechanics

#### 1. Case 1: Burgers equation

The Burgers equation is a significant PDE employed to simulate the propagation and reflection of waves in various fields, including fluid mechanics, nonlinear acoustics, and gas dynamics.[33] The general form of the Burgers equation is as follows:

$$\mathbf{u}_t = -\mathbf{u} \cdot \mathbf{\nabla u} + \upsilon\Delta\mathbf{u} \tag{19}$$

where $\mathbf{u}$ is the velocity vector, $\upsilon$ denotes the viscosity, and $\Delta$ denotes the Laplace operator.

In this case, we consider the two-dimensional Burgers equation, $\mathbf{u} = (u, v)^T$, and viscosity $\upsilon = 0.1$. The datasets are generated with periodic boundary conditions on the domain of $\Omega = [-\pi, \pi] \times [-\pi, \pi]$ with 256×256 identical mesh on the time domain $t \in [0,4]$ with $\delta t = 0.01$. To simulate the evolution of shock wave, the following initial condition is employed here[34]

$$\mathbf{u}_0 = A_1 \pm B_1 \times \text{sech}(C_1 \times ((x \pm D_1)^2 + (y \pm D_1)^2)) \tag{20}$$

where $A_1$, $B_1$, $C_1$, and $D_1$ are constant, in which, we set $A_1 = 0$, $B_1 = 8$, $C_1 = 4$, and $D_1 = 1$.

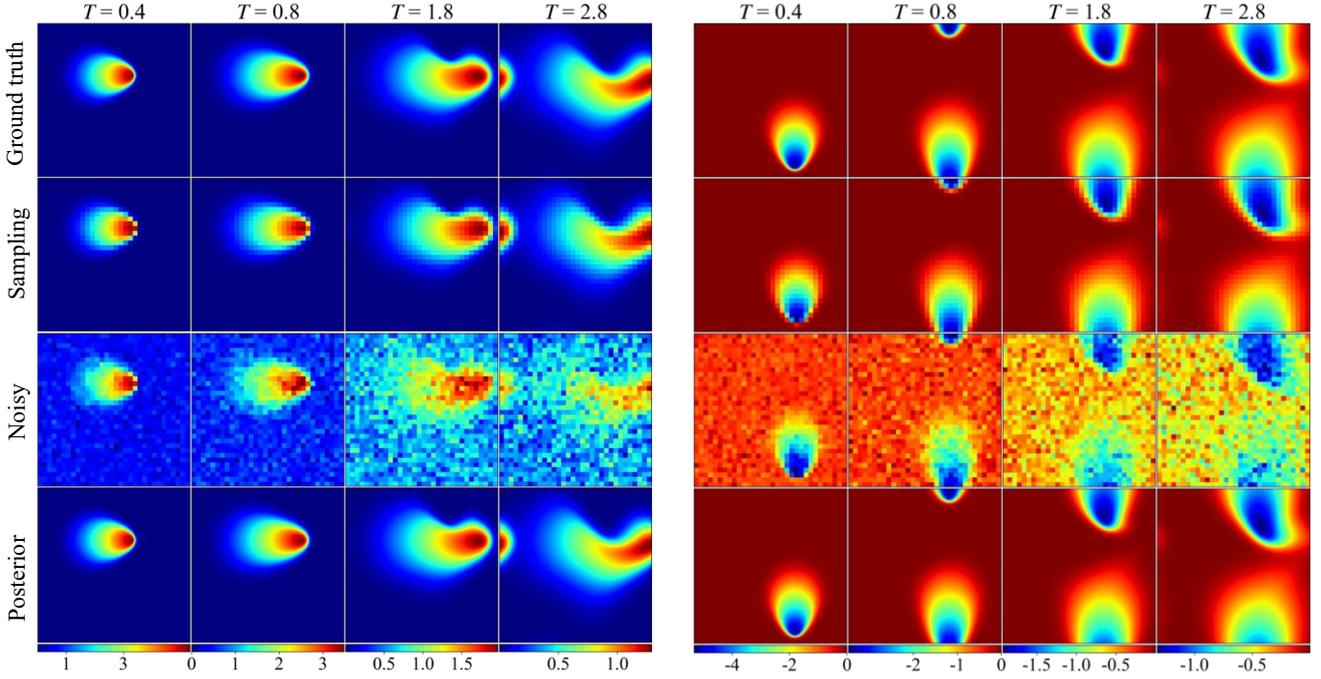

**FIG. 2.** Traning data snapshots and posterior value snapshots of $u$ and $v$ for Burgers equation over two-dimensional domain $\Omega$, left is $u$ component and right is the $v$ component.

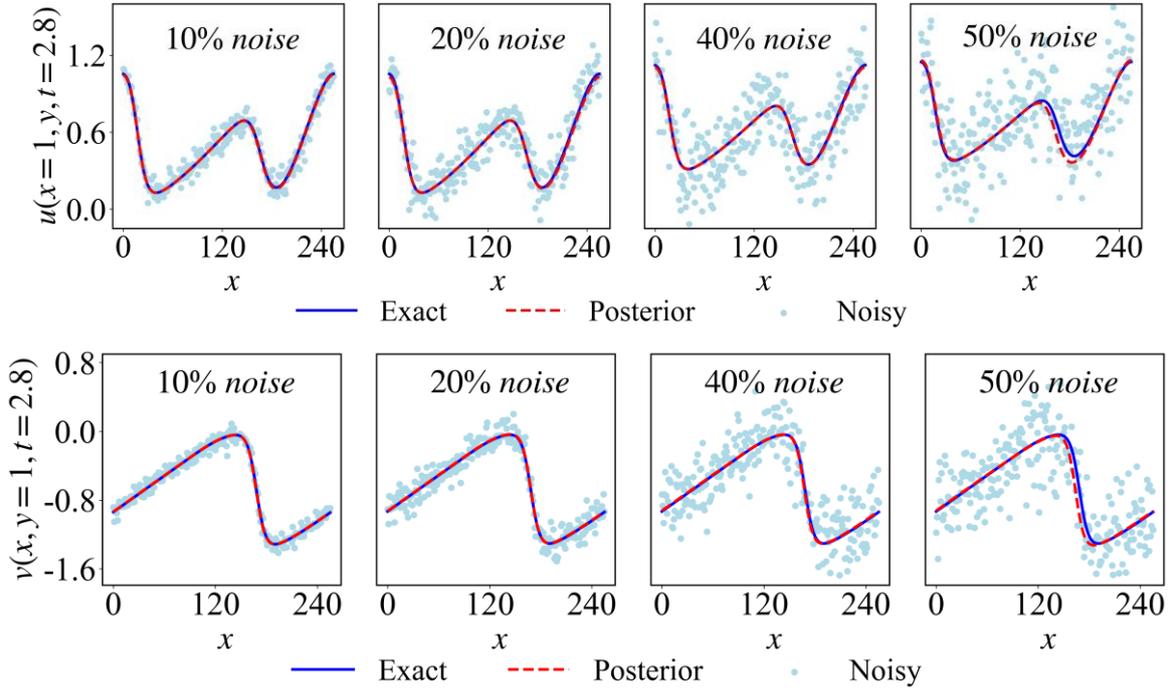

**FIG. 3** Comparison among Exact data, Noisy data, and posterior data for the Burgers equation at different levels of noise.

To simulate a more realistic scenario of equation discovery from low-quality datasets, we downsample the data from a 256×256 mesh to a 32×32 mesh in space, as illustrated in Fig. 2. Gaussian white noise is added to the sparse data in Fig. 2. During the training process, 30 time-step data points, randomly selected from 120 continuous time steps, are utilized to train the neural network. A deep neural network with 10 hidden layers is constructed and per hidden layer has 60 neurons. The activation function of neural network employed here is tanh( • ). In the loss function, the weight coefficient $\alpha$ is set to be 1. $\beta$ is initially set to be $10^{-7}$ and gradually increased to $10^{-6}$ during the training process until the L-BFGS-B optimizer converges. Simultaneously, the size of the candidate library remains unchanged. In this example, the Adam optimizer is employed for 2000 epochs, followed by the use of L-BFGS-B to expedite the training. The L-BFGS-B automatically stops when the training converges. The candidate library used by PDE-FIND[7] and PiDL[16] in this context is as follows, based on the construction principle in[16], comprising 110 candidates in the library $\Theta$.

$$\Theta = [1, u_x, u_y, \dots, v_{xy}, v_{yy}, v, u, \dots, u^2v, u^3, vu_x, uu_x, \cdots, u^2vv_{yy}, u^3v_{yy}] \tag{21}$$

However, the library embedded with Galilean invariance contains 15 candidates, thus preventing the inclusion of numerous miscellaneous and meaningless terms.

$$\Theta_G = [1, uu_x, vu_y, uv_x, vv_y, u_x, u_y, u_{xx}, u_{xy}, u_{yy}, v_x, v_y, v_{xx}, v_{xy}, v_{yy}] \tag{22}$$

Eq. (22) comprises the detailed terms employed by ICNet, satisfying the requirement of Galilean invariance. ICNet with STRidge is used to obtain parsimonious equations. Since ICNet has discovered the correct equation, we can directly input the coefficients obtained from data into numerical simulation to assess the error compared with the true solution. Fig. 3 illustrates the matching degree of posterior data and true solution at different levels of noise. It can be observed from Fig. 3 that ICNet can accurately capture the shock behavior even from sparse data with high noise. Additionally, Fig. 4 demonstrates that after the training with the Adam optimizer, the L-BFGS-B optimizer can rapidly discover the true coefficients and maintain their stability. Fig. 4 presents the numerical experimental results on sparse data with 50% noise.

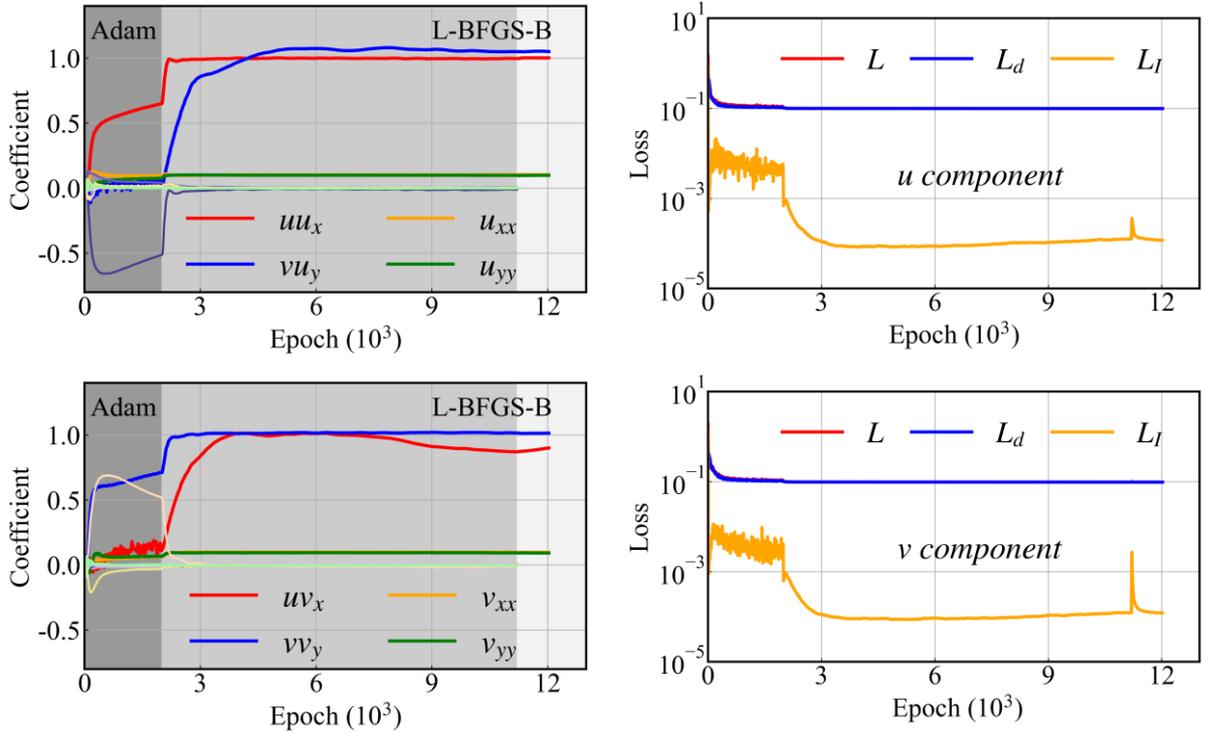

**FIG. 4.** Left is the coefficients variation along with the number of epochs for Burgers equation. The third part of the shaded area in the figure illustrates that $\|\Lambda\|_0$ remains unchanged. Except for the legends annotated in the figure, the remaining lines represent variations of other candidate terms. The right is the loss function of training process.

Next, we compare ICNet with existing methods, starting with a comparison between ICNet and PDE-FIND on sparse data with varying levels of noise. The comparison results are summarized in Table 1. All comparison tests, except for the first line in Table 1, are conducted on 32×32 sparse data. As seen in Table 1, both ICNet and PDE-FIND can identify accurate equations with densely distributed data (256×256) and without noise. However, ICNet demonstrates superior accuracy and robustness when dealing with sparse and noisy data.

TABLE I. Discovery of Burgers equation by ICNet and PDE-FIND with different levels of noise.

| | Correct PDE | $u_t = -uu_x - vu_y + 0.1(u_{xx} + u_{yy})$ $v_t = -uv_x - vv_y + 0.1(v_{xx} + v_{yy})$ | |
|---|---|---|---|
| Noise | | ICNet | PDE-FIND |
| 0% (256×256) | | $u_t = -1.00uu_x - 1.00vu_y + 0.100u_{xx}$ $+0.100u_{yy}$ $v_t = -0.998uv_x - 1.00vv_y + 0.100v_{xx}$ $+0.100v_{yy}$ | $u_t = -0.999uu_x - 1.00vu_y + 0.100u_{xx}$ $+0.101u_{yy}$ $v_t = -1.00uv_x - 1.00vv_y + 0.101v_{xx}$ $+0.100v_{yy}$ |
| 0% | | $u_t = -1.00uu_x - 0.999vu_y + 0.100u_{xx}$ $+0.100u_{yy}$ $v_t = -0.999uv_x - 1.00vv_y + 0.100v_{xx}$ $+0.100v_{yy}$ | $u_t = -0.703uu_x - 0.879vu_y + 0.115u_{xx}$ $+0.060u_{yy} - \mathbf{0.580}u^2u_y \ldots \ldots$ $v_t = -0.936uv_x - 1.060vv_y + 0.067v_{xx}$ $+0.109v_{yy} - \mathbf{0.194}v^3v_x \ldots \ldots$ |
| 10% | | $u_t = -0.999uu_x - 1.023vu_y + 0.100u_{xx}$ $+0.100u_{yy}$ $v_t = -1.051uv_x - 1.003vv_y + 0.100v_{xx}$ $+0.099v_{yy}$ | $u_t = -1.052uu_x - 0.00vu_y + 0.070u_{xx}$ $+0.052u_{yy} - \mathbf{4.439}uvu_x \ldots \ldots$ $v_t = -0.00uv_x - 0.774vv_y + 0.00v_{xx}$ $+0.066v_{yy} - \mathbf{0.618}v^2 \ldots \ldots$ |

TABLE II. Discovery of Burgers equation by ICNet and PiDL with higher levels of noise.

| | Correct PDE | $u_t = -uu_x - vu_y + 0.1(u_{xx} + u_{yy})$ $v_t = -uv_x - vv_y + 0.1(v_{xx} + v_{yy})$ | |
|---|---|---|---|
| Noise | | ICNet | PiDL |
| 30% | | $u_t = -0.998uu_x - 1.05vu_y + 0.101u_{xx}$ $+0.100u_{yy}$ $v_t = -1.08uv_x - 1.01vv_y + 0.100v_{xx}$ $+0.097v_{yy}$ | $u_t = -0.395uu_x - 0.922vu_y + 0.098u_{xx}$ $+0.083u_{yy} - 0.385u_x - \mathbf{0.244}u^2u_x \ldots \ldots$ $v_t = -0.00uv_x - 0.765vv_y + 0.081v_{xx}$ $+0.099v_{yy} + 0.160v_y + \mathbf{0.099}v^2v_y \ldots \ldots$ |
| 40% | | $u_t = -0.999uu_x - 1.03vu_y + 0.102u_{xx}$ $+0.099u_{yy}$ $v_t = -1.02uv_x - 1.01vv_y + 0.099v_{xx}$ $+0.095v_{yy}$ | $u_t = -0.611uu_x - 0.884vu_y + 0.109u_{xx}$ $+0.089u_{yy} - 0.278u_x - \mathbf{0.130}u^2u_x \ldots \ldots$ $v_t = -0.00uv_x - 0.950vv_y + 0.089v_{xx}$ $+0.088v_{yy}$ |
| 50% | | $u_t = -1.001uu_x - 1.050vu_y + 0.103u_{xx}$ $+0.098u_{yy}$ $v_t = -0.899uv_x - 1.013vv_y + 0.098v_{xx}$ $+0.092v_{yy}$ | $u_t = -0.937uu_x - 0.00vu_y + 0.099u_{xx}$ $+0.087u_{yy}$ $v_t = -0.00uv_x - 0.856vv_y + 0.053v_{xx}$ $+0.076v_{yy} + \mathbf{0.553}v \ldots \ldots$ |

Subsequently, we compare ICNet with the PiDL method in this study. We examine the impact of higher noise levels (30%, 40%, and 50% noise added to sparse data). The results of the two methods are presented in Table 2. Notably, even with the use of automatic differentiation, PiDL still fails to correctly discover results from datasets with high levels of noise. Nevertheless, the ICNet can discover the correct PDE with higher accuracy, indicating that the ICNet taking account of invariance is more robust than the existing methods. Additionally, beyond the accuracy of discovered equations, the relative error of the model serves as an indicator to assess the performance of the two methods. Since PiDL does not identify the true PDE, the trained neural network is employed to generate datasets $\mathbf{u}$ at the training time-steps. The relative error $\varepsilon_t$ is then calculated,

$$\varepsilon_t = \frac{\|\hat{\mathbf{u}}(x,y,t) - \mathbf{u}(x,y,t)\|_2^2}{\|\mathbf{u}(x,y,t) - \tilde{\mathbf{u}}(x,y,t)\|_2^2} \tag{23}$$

where $\hat{\mathbf{u}}(x,y,t)$ is the predicted data using neural network, $\mathbf{u}(x,y,t)$ is the true data and $\tilde{\mathbf{u}}(x,y,t)$ is the spatial average value of true data at $t$ moment.

Fig. 5 illustrates the evolution of relative errors for ICNet and PiDL. The superior performance of ICNet is evident, affirming the effectiveness of candidates based on invariance.

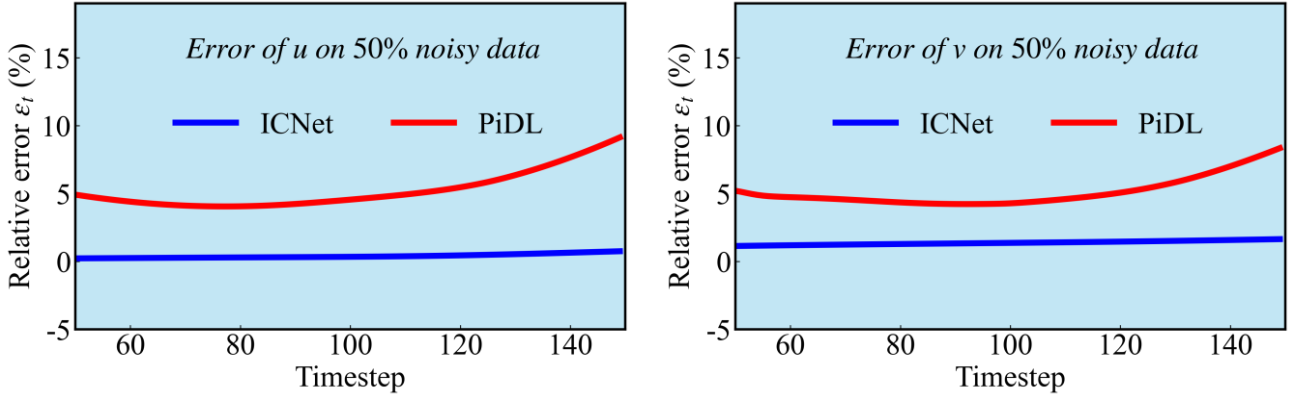

**FIG. 5.** Relative error $\varepsilon_t$ of ICNet and PiDL.

## 2. Case 2: Equation of Stenotic 2D channel flow over an obstacle

The Navier-Stokes (N-S) equation is discovered from the dataset of 2D channel flow over an obstacle body with a diameter of 10, as shown in Fig. 6. The channel flow datasets come from literature.[29] The governing equations of this flow are the incompressible Navier-Stokes equations, given by:

$$\partial_t \mathbf{u} = -(\mathbf{u} \cdot \boldsymbol{\nabla})\mathbf{u} - \nabla p + Re^{-1}\Delta \mathbf{u}$$
$$\boldsymbol{\nabla} \cdot \mathbf{u} = 0 \qquad (24)$$

where the $Re$ denotes the Reynolds number, here $Re = 5$, $\Delta$ denotes the Laplace operator, $\mathbf{u} = (u,v)^T$ is the nondimensional velocity vector and $p$ is nondimensional pressure. The spatial domain of the channel flow is $[15,55] \times [0,12]$ and time interval is $[0,20]$ with time step $\delta t = 0.1$.

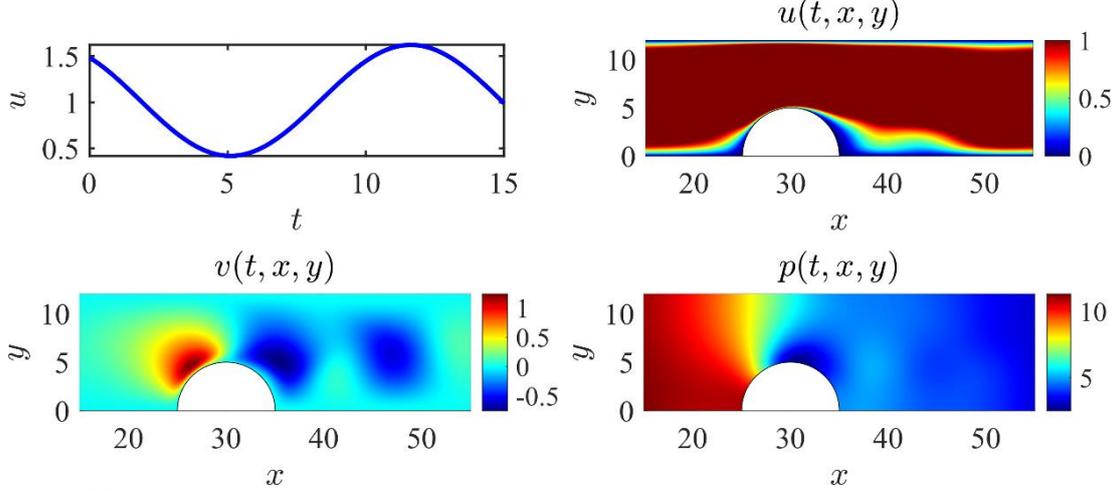

**FIG. 6.** Left of first row is the velocity profile imposed at the inlet. The remaining three sub-graphs is velocity and pressure snapshots at $t = 5.5$.

Sixty continuous time-step data points are used to train the neural network and discover equations. In this study, A deep neural network with 10 hidden layers is constructed and per hidden layer has 60 neurons. The activation function of neural network employed here is tanh( · ). In the loss function, the weight coefficient $\alpha$ is set to be 1, and the $\beta$ is adjusted to increase gradually until the coefficients that have been discovered stably are suppressed. During the training process, the Adam optimizer is used to train the neural network with 40,000 epochs and $\beta = 10^{-7}$ initially. If the $\beta$ is set to be large at the beginning, the true coefficients value of active terms would be suppressed. Then, the $\beta$ is set to be $10^{-5}, 10^{-4}, 10^{-3}, 10^{-3}, 10^{-3}, 10^{-3}$, and the neural network is trained with 40,000 epochs, 20,000 epochs, 20,000 epochs, 20,000 epochs, 20,000 epochs, 20,000 epochs by Adam optimizer respectively. We train the neural network until the size of candidates remains unchanged. The ICNet, integrated with STRidge during the training process without fixing learnable parameters, can further decrease the equation residuals, as demonstrated in Appendix A. The variation of coefficients during the training process is shown in Fig. 7.

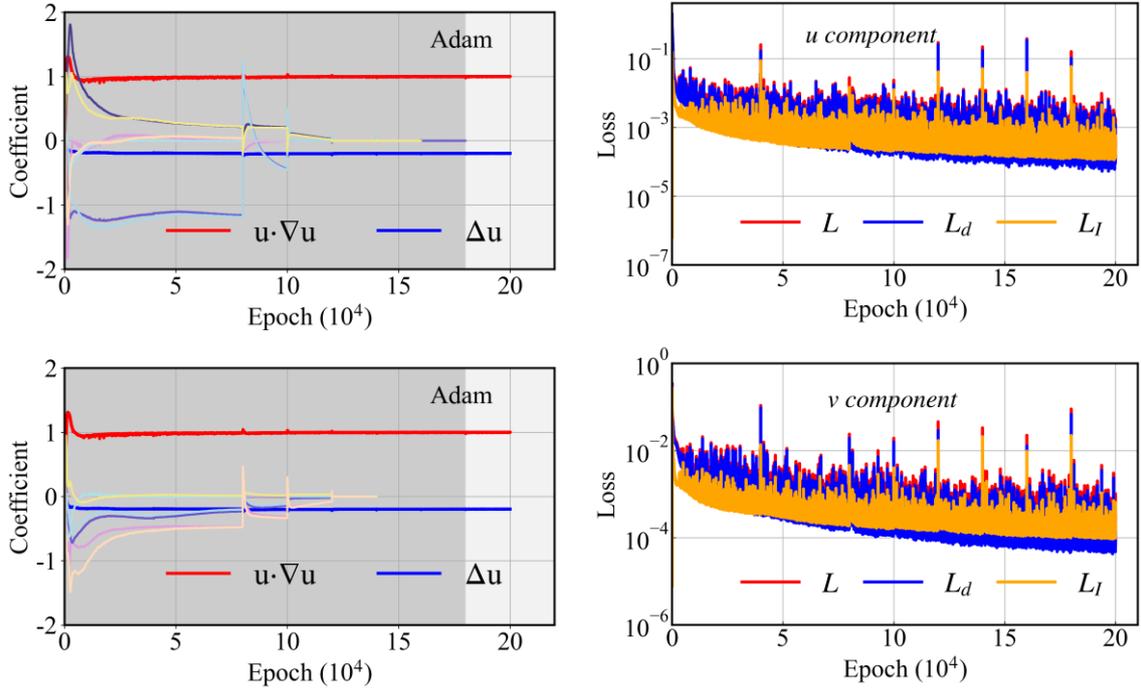

**FIG. 7.** Left is the coefficients variation along with the number of epochs for 2D Channel flow over an obstacle. The second part of the shaded area in the figure illustrates that $\|\mathbf{\Lambda}\|_\mathbf{0}$ remains unchanged. Except for the legends annotated in the figure, the remaining lines represent variations of other candidate terms. The right is the loss function of training process.

The library embedded with invariance is as follows,
$$\Theta_G = [1, uu_x, vu_y, uv_x, vv_y, u_x, u_y, u_{xx}, u_{xy}, u_{yy}, v_x, v_y, v_{xx}, v_{xy}, v_{yy}] \tag{25}$$

Table 3 and Table 4 show the equations discovered by ICNet compared with PDE-FIND and PiDL. Besides the worse accuracy of coefficients, it should be noted that there are still many other redundant terms with large coefficients discovered by PDE-FIND compared with the true PDE. To verify that the existence of redundant terms is unreasonable, we compared the equation residuals (eqrs) achieved by ICNet and PiDL, as shown in Fig. 8. It can be observed that the equations discovered by ICNet exhibit smaller equation residuals.

**TABLE III.** N-S equation discovered from 2D channel flow dataset by ICNet and PDE-FIND

| | Correct PDE | $u_t = -uu_x - vu_y - p_x + 0.2(u_{xx} + u_{yy})$ $v_t = -uv_x - vv_y - p_y + 0.2(v_{xx} + v_{yy})$ | |
|---|---|---|---|
| | | ICNet | PDE-FIND (DL-PDE) |
| Results | | $u_t = -0.999uu_x - 0.999vu_y - 1.00p_x$ $+0.195u_{xx} + 0.195u_{yy}$ $v_t = -0.999uv_x - 0.999vv_y - 1.00p_y$ $+ 0.202v_{xx} + 0.201v_{yy}$ | $u_t = -0.972uu_x - 1.030vu_y - 0.898p_x$ $+0.487u_x + 0.168u_{xx} + 0.084u_{yy} + \mathbf{0.124}u^2v_y \ldots$ $v_t = -0.864uv_x - 0.829vv_y - 0.795p_y$ $+0.348u_x + 0.132v_{xx} + 0.029v_{yy} + \mathbf{0.232}vv_x \ldots$ |

**TABLE IV.** N-S equation discovered from 2D channel flow dataset by ICNet and PiDL

| | Correct PDE | $u_t = -uu_x - vu_y - p_x + 0.2(u_{xx} + u_{yy})$ $v_t = -uv_x - vv_y - p_y + 0.2(v_{xx} + v_{yy})$ | |
|---|---|---|---|
| | | ICNet | PiDL |
| Results | | $u_t = -0.999uu_x - 0.999vu_y - 1.00p_x$ | $u_t = -0.923uu_x - 0.982vu_y - 1.00p_x$ |

$$v_t = -0.999uv_x - 0.999vv_y - 1.00p_y$$
$$+0.195u_{xx} + 0.195u_{yy}$$
$$+0.202v_{xx} + 0.201v_{yy}$$

$$+0.848v_y + 0.816u_x - 0.292v_{yy}$$
$$-0.277u_{xy} + \mathbf{0.094}uv_y \dots$$
$$v_t = -0.973uv_x - 0.668vv_y - 1.00p_y$$
$$+\mathbf{0.999}v^3u_x + \mathbf{0.983}v^3v_y + 0.400v_{xy}$$
$$+0.363u_{xx} \dots$$

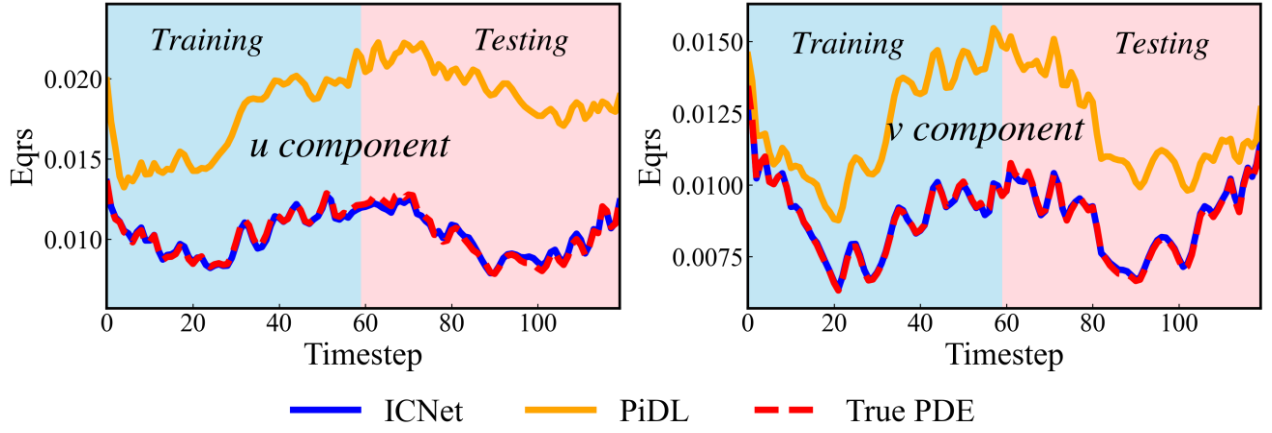

**FIG. 8.** Equation residuals achieved by ICNet and PiDL. Left is $u$ component and right is $v$ component.

### 3. Case 3: Equation of 3D intracranial aneurysm

The Navier-Stokes (N-S) equation discovered from the 3D intracranial aneurysm, as shown in Fig. 9, is performed. The datasets for this example are sourced from.[29] The governing equations of this incompressible Newtonian fluid are the 3D Navier-Stokes equations as follows,

$$\partial_t \mathbf{u} = -(\mathbf{u} \cdot \nabla)\mathbf{u} - \nabla p + Re^{-1}\Delta \mathbf{u}$$
$$\nabla \cdot \mathbf{u} = 0 \tag{26}$$

where the Reynolds number $Re = 98.2$, $\Delta$ denotes the Laplace operator, $\mathbf{u} = (u, v, w)^T$ is non-dimensional velocity vector and $p$ is non-dimensional pressure.

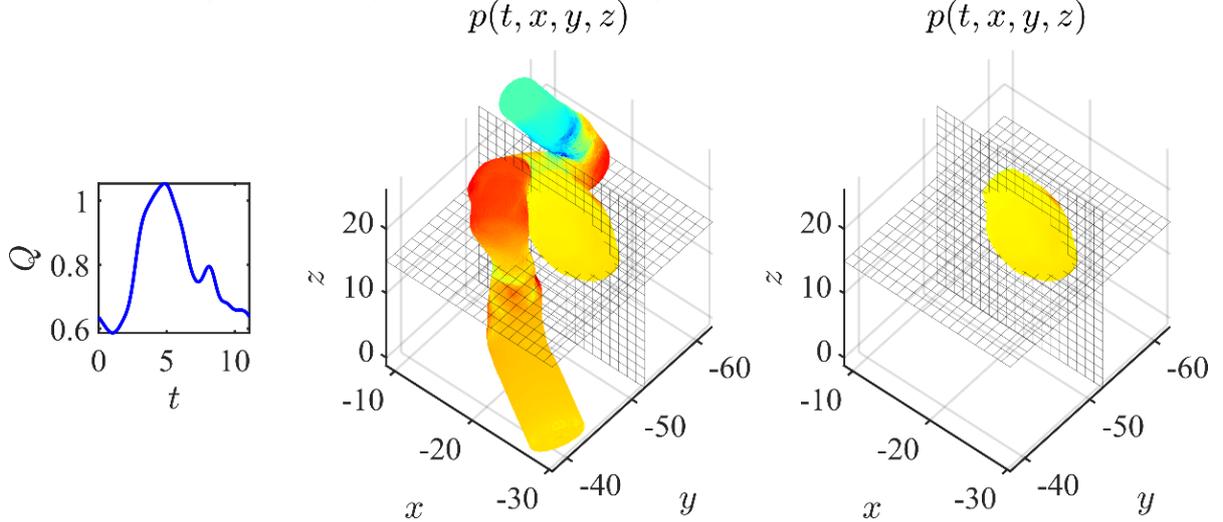

**FIG. 9.** Flow field of 3D intracranial aneurysm, the aneurysm attached to an artery is the research domain in this study, as shown in right. The left is the waveform of physiologic flow $Q$ at the inlet; The middle is the pressure field of intracranial aneurysm; The right is the pressure field of the aneurysm.[29]

Sixty continuous time-step data points with $\delta t = 0.1$ are used for training the neural network. The entire spatial domain comprises 689,391 data points. In this study, A deep neural network with 10 hidden layers is constructed and per hidden layer has 150 neurons. The activation function of neural network employed here is tanh( · ). The weight coefficient $\alpha$ is set to 0.5 in this scenario. Initially, the Adam optimizer is utilized to train the neural network for 140,000 epochs with a small $\beta$ value of $10^{-7}$ to learn the coefficients. Subsequently, the $\beta$ value is gradually adjusted to $10^{-5}, 10^{-4}, 10^{-3}, 5 \times 10^{-3}, 5 \times 10^{-3}, 5 \times 10^{-3}$, and the neural network is trained for 70,000 epochs, 70,000 epochs, 40,000 epochs, 20,000 epochs, and 20,000 epochs, respectively. Fig. 10 illustrates the variation of coefficients during the training process. The effectiveness of ICNet with STRidge in further reducing equation residuals is confirmed, as demonstrated in Appendix A.

**TABLE V.** N-S equation discovered from 3D intracranial aneurysm by ICNet and PDE-FIND (DL-PDE)

| Correct PDE | $u_t = -uu_x - vu_y - wu_z + p_x + 0.0102(u_{xx} + u_{yy} + u_{zz})$ <br> $v_t = -uv_x - vv_y - wv_z + p_y + 0.0102(v_{xx} + v_{yy} + v_{zz})$ <br> $w_t = -uw_x - vw_y - ww_z + p_z + 0.0102(w_{xx} + w_{yy} + w_{zz})$ | |
|---|---|---|
| | ICNet | PDE-FIND (DL-PDE) |
| Results | $u_t = -0.955uu_x - 0.955vu_y - 0.955wu_z$ <br> $+1.00p_x + 0.0100u_{xx} + 0.0110u_{yy}$ <br> $+0.0097u_{zz}$ <br> $v_t = -0.955uv_x - 0.955vv_y - 0.955wv_z$ <br> $+1.00p_y + 0.0097v_{xx} + 0.0107v_{yy}$ <br> $+0.0097v_{zz}$ <br> $w_t = -0.955uw_x - 0.955vw_y - 0.955ww_z$ <br> $+1.00p_z + 0.0100w_{xx} + 0.0099w_{yy}$ <br> $+0.0114w_{zz}$ | $u_t = -0.162uu_x - 0.109vu_y - 0.146wu_z$ <br> $+0.078p_x + 0.0111u_x + 0.00u_{xx}$ <br> $+0.00u_{yy} + 0.00u_{zz}$ <br> $v_t = -0.106uv_x - 0.080vv_y - 0.127wv_z$ <br> $+0.049p_y - 0.0124v_x + 0.00v_{xx}$ <br> $+0.0006v_{yy} + 0.00v_{zz}$ <br> $w_t = -0.079uw_x - 0.059vw_y - 0.00ww_z$ <br> $+0.00p_z + 0.00w_{xx} + 0.00w_{yy}$ <br> $+0.00w_{zz}$ |

**TABLE VI.** N-S equation discovered from 3D intracranial aneurysm by ICNet and PiDL

| Correct PDE | $u_t = -uu_x - vu_y - wu_z + p_x + 0.0102(u_{xx} + u_{yy} + u_{zz})$ <br> $v_t = -uv_x - vv_y - wv_z + p_y + 0.0102(v_{xx} + v_{yy} + v_{zz})$ <br> $w_t = -uw_x - vw_y - ww_z + p_z + 0.0102(w_{xx} + w_{yy} + w_{zz})$ | |
|---|---|---|
| | ICNet | PiDL |
| Results | $u_t = -0.955uu_x - 0.955vu_y - 0.955wu_z$ <br> $+1.00p_x + 0.0100u_{xx} + 0.0110u_{yy}$ <br> $+0.0097u_{zz}$ <br> $v_t = -0.955uv_x - 0.955vv_y - 0.955wv_z$ <br> $+1.00p_y + 0.0097v_{xx} + 0.0107v_{yy}$ <br> $+0.0097v_{zz}$ <br> $w_t = -0.955uw_x - 0.955vw_y - 0.955ww_z$ <br> $+1.00p_z + 0.0100w_{xx} + 0.0099w_{yy}$ <br> $+0.0114w_{zz}$ | $u_t = -0.00uu_x - 0.429vu_y - 0.00wu_z$ <br> $+1.00p_x - 0.00u_{xx} - 0.12u_{yy} - 0.00u_{zz}$ <br> $+\mathbf{1.185}u^2u_z - \mathbf{1.113}uu_x + \mathbf{0.721}uv_y$ <br> $+\mathbf{0.682}u^2v_y + \mathbf{0.543}wu_x + \mathbf{0.465}u^2u_x \ldots \ldots$ <br> $v_t = -0.865uv_x - 0.380vv_y - 0.00wv_z$ <br> $+1.00p_y - 0.032v_{xx} - 0.00v_{yy} + 0.00v_{zz}$ <br> $+\mathbf{0.718}u^2v_x + \mathbf{0.567}uv_{xz} + 0.478v_x$ <br> $+\mathbf{0.329}vw_x - \mathbf{0.343}u^2v_z + \mathbf{0.163}v \ldots \ldots$ <br> $w_t = -1.089uw_x - 0.509vw_y - 0.00ww_z$ <br> $+1.00p_z - 0.00w_{xx} - 0.00w_{yy} - 0.00w_{zz}$ <br> $+\mathbf{0.599}u^2w_x + \mathbf{0.490}wu_x - \mathbf{0.650}uwu_y$ <br> $-0.331w_x + \mathbf{0.104}uw + \mathbf{0.074}uv_{zz} \ldots \ldots$ |

The library embedded with invariance in this example is as follows:

$$\Theta_G = [1, uu_x, vu_y, wu_z, \ldots, uw_x, vw_y, ww_z, u_x, u_y, u_z, \ldots w_x, w_y, w_z, u_{xx}, u_{xy}, \ldots, w_{xz}, w_{yz}] \quad (27)$$

The above library consists of only 32 candidates, whereas the over-complete library, following the construction principle in,[10] contains 560 candidates. The use of a large library could escalate the challenge of sparse regression,[10] demanding more computational resources and time for extensive data. Hence, the benefits of the invariance library are particularly notable in solving three-dimensional complex fluid problems. Table 5 presents the equations discovered by ICNet and PDE-FIND. It is evident that PDE-FIND not only retrieves inferior coefficient values but also fails to identify the correct terms of the PDE.

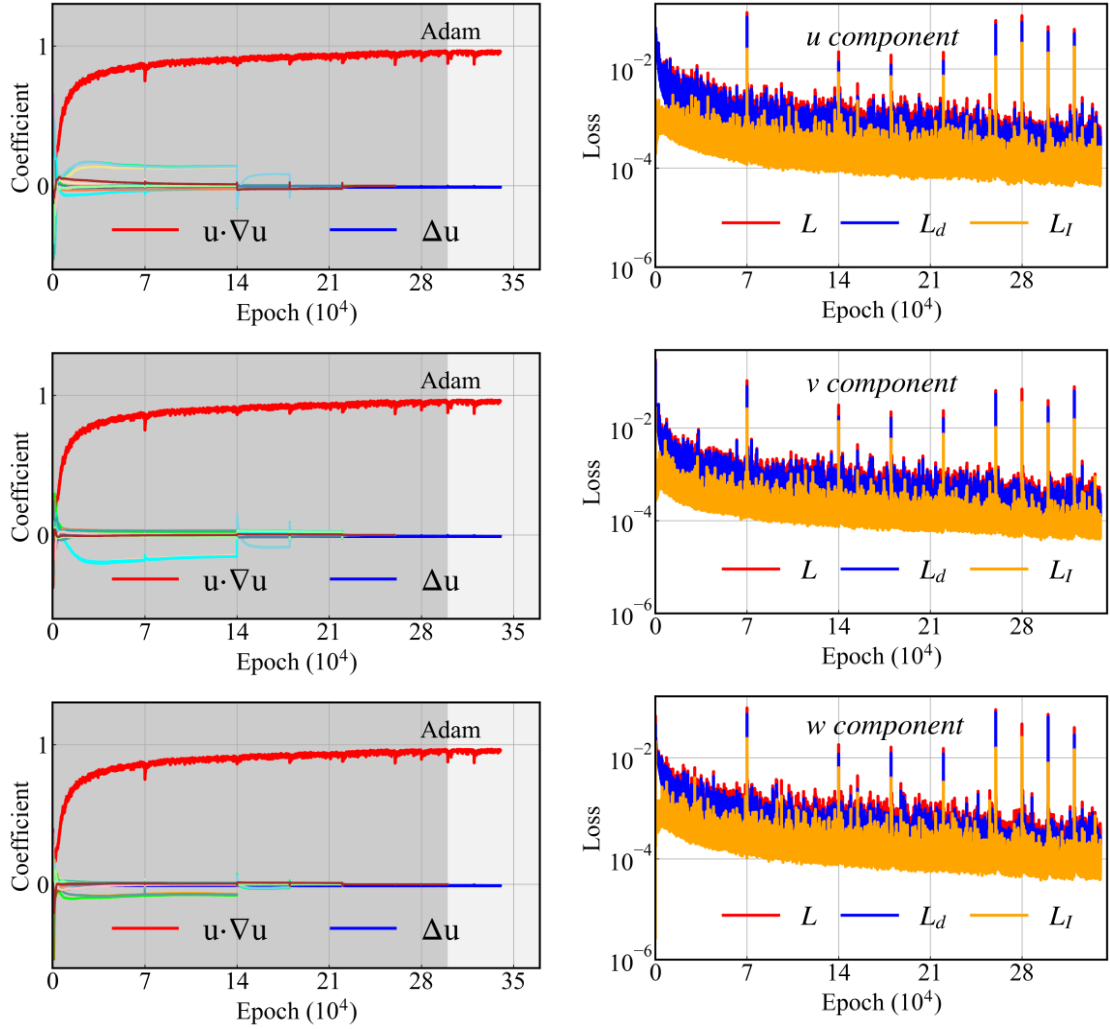

**FIG. 10.** Left is the coefficients variation along with the number of epochs for 3D intracranial aneurysm. The second part of the shaded area in the figure illustrates that $\|\Lambda\|_0$ remains unchanged. Except for the legends annotated in the figure, the remaining lines represent variations of other candidate terms. The right is the loss function of training process.

The equations discovered by ICNet and PiDL are presented in Table 6. The table indicates that PiDL discovers equations with lower accuracy compared to ICNet and fails to identify the true terms of PDEs. The comparison of equation residuals for ICNet and PiDL is illustrated in Fig. 11. The comparison highlights that both PDE-FIND and PiDL fails to obtain parsimonious and reasonable results for such complex fluids, even when utilizing clean data.

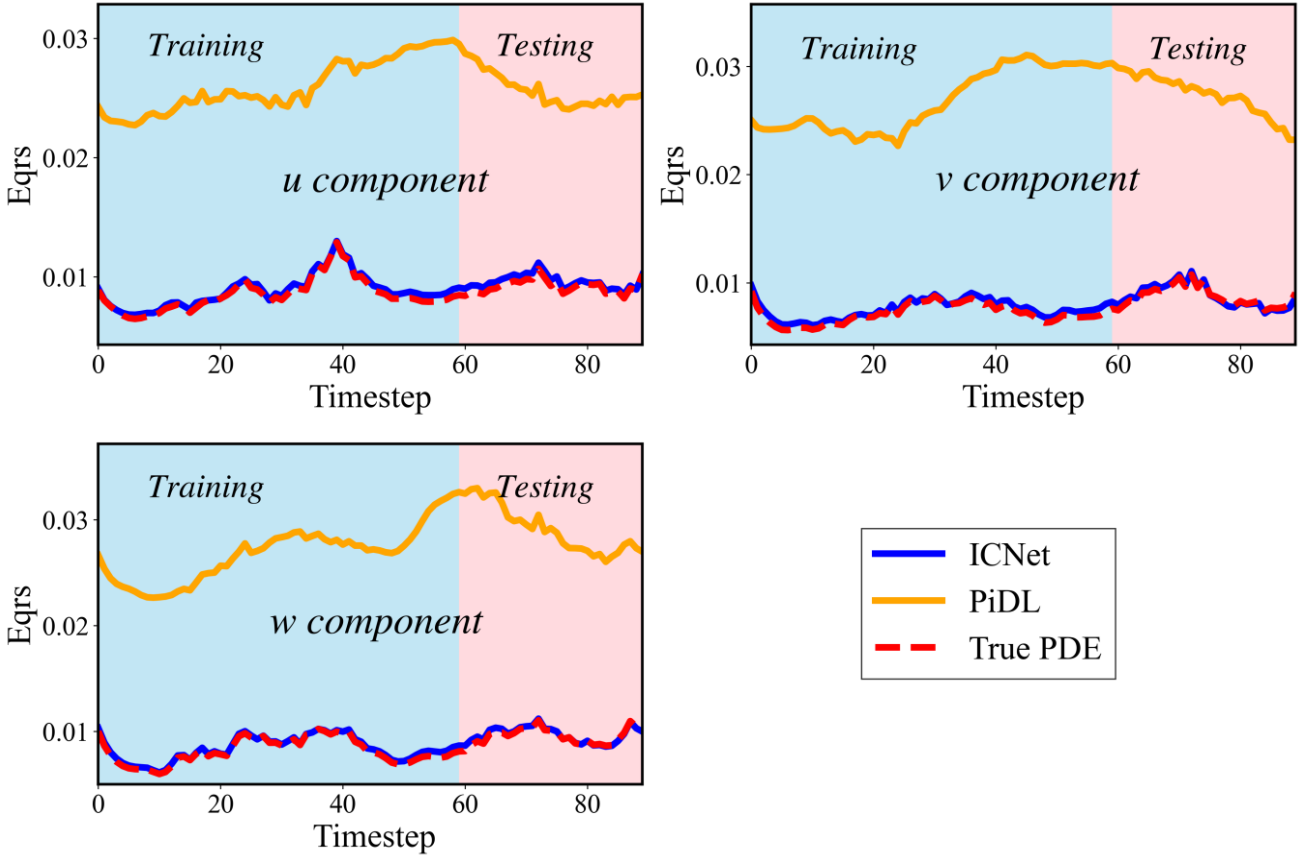

FIG. 11. Equation residuals comparison of ICNet and PiDL.

## B. Numerical examples for relativity

### 1. Case 4: Single Klein-Gordon equation

In theoretical physics, scalar field theory plays a crucial role in describing significant physical phenomena in the fields of particle physics, astrophysics, and cosmology.[35] The classical scalar field equation with Lorentz invariance is the nonlinear Klein-Gordon equation, represented as follows:

$$\phi_{tt} = a_1 \phi + b_1 \phi^3 + d_1 \Delta \phi \tag{28}$$

where $\phi$ is the scalar of single scalar field, $a_1$, $b_1$ and $d_1$ are constant, and $\Delta$ denotes the Laplace operator. In this case, we take the $a_1 = 1$, $b_1 = -1$ and $d_1 = 0.1$ to generate the data on the two-dimensional domain of $\Omega = [-\pi, \pi] \times [-\pi, \pi]$ with periodic boundary conditions. We set $\delta t = 0.01$ during the process of simulation with time domain $t \in [0,4]$. The exponential initial condition is employed to simulate the evolution of the scalar field.[36,37]

$$\phi_0 = A_2 \times exp\bigl(-B_2((x - x_0)^2 + (y - y_0)^2)\bigr) \tag{29}$$

where we set $A_2 = 4$, $B_2 = 10$, $x_0 = 0$, and $y_0 = 0$.

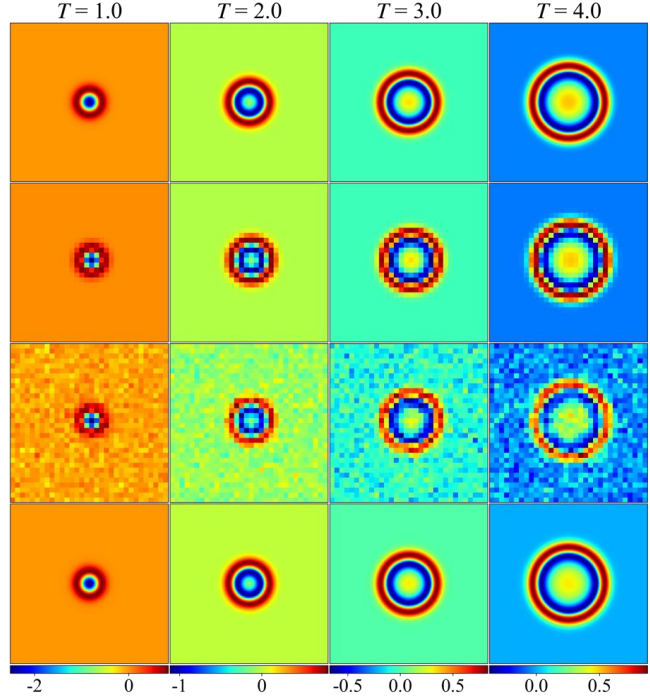

**FIG. 12.** Training data snapshots and posterior data snapshots of $\phi$ for Single Klein-Gordon equation over two-dimensional domain $\Omega$.

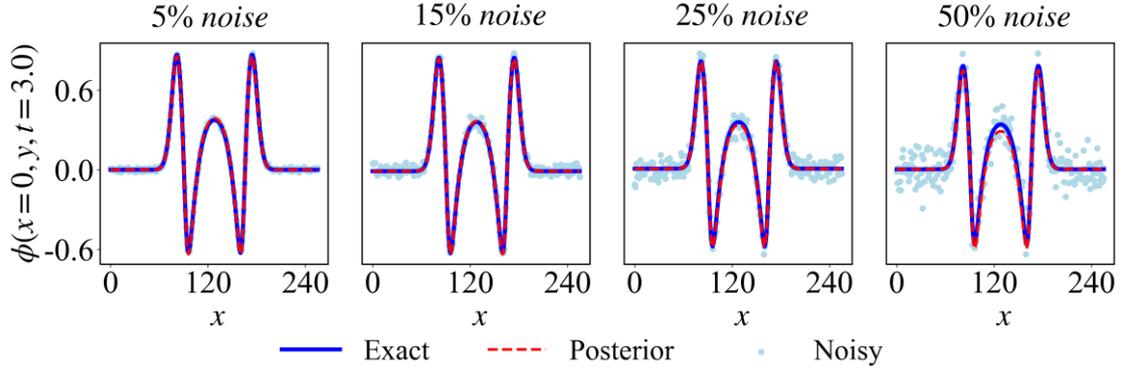

**FIG. 13.** Comparison of Exact data, Noisy data, and posterior data for the Single Klein-Gordon equation at different level noise.

We also downsample the simulation data from 256×256 mesh to 32×32 mesh, and then the downsampling data is used to train the neural network. The most intensive noise of 50% is added to the sparse data, as illustrated in Fig. 12. A deep neural network with 8 hidden layers is constructed and per hidden layer has 30 neurons. The activation function of neural network employed here is tanh( • ). In the loss function, the weight coefficient $\alpha$ is set to 1, and $\beta$ is set to $10^{-7}$ and $10^{-6}$ during the training process until the L-BFGS-B optimizer converges. Meanwhile, the size of candidates remains unchanged. The ICNet is trained for 2000 iterations using Adam optimizer, and then the L-BFGS-B optimizer is employed for acceleration. Fig. 13 illustrates the excellent match between posterior data and the true solution at different levels of noise, even with limited data. Fig. 14 displays the variation of coefficients and the loss function during the training process for the Single Klein-Gordon equation.

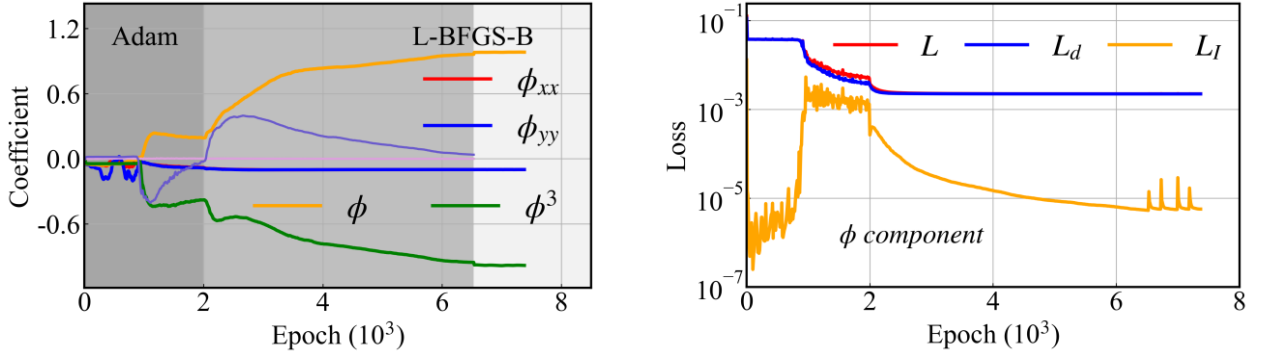

**FIG. 14.** Left is the variation of coefficients along with the number of epochs for Single Klein-Gordon equation. The third part of the shaded area in the figure illustrates that $\|\Lambda\|_0$ remains unchanged. Except for the legends annotated in the figure, the remaining lines represent variations of other candidate terms. The right is the evolution of loss function.

We then compare ICNet with PDE-FIND and PiDL. The comparison results of ICNet and PDE-FIND at different levels of noise are shown in Table 7. Table 7 reveals that ICNet can provide very accurate coefficients even with high noise and sparse data, while PDE-FIND can achieve correct results only when the quality and quantity of data are higher. Table 8 presents the comparison results of ICNet and PiDL with higher noise. Table 8 demonstrates the advantages of the proposed method embedded with Lorentz invariance. Fig. 15 shows the superior performance over the training data of ICNet compared with PiDL at each time step $\delta t = 0.01$. There are the following 24 candidates used in PDE-FIND,

$$\Theta = [1, \phi_x, \phi_y, \ldots, \phi_{xx}, \phi_{yy}, \phi, \phi^2, \phi^3, \phi\phi_x, \phi^2\phi_x, \cdots, \phi^2\phi_{yy}, \phi^3\phi_{yy}] \tag{30}$$

The library embedded with Lorentz invariance is as follows,

$$\Theta_L = [1, \phi_{xx}, \phi_{yy}, \phi, \phi^2, \phi^3] \tag{31}$$

**TABLE VII.** Discovery of Single Klein-Gordon equation by ICNet and PDE-FIND with different levels of noise.

| | Correct PDE $\phi_{tt} = \phi - \phi^3 + 0.1(\phi_{xx} + \phi_{yy})$ | |
|---|---|---|
| Noise | ICNet | PDE-FIND |
| 0% (256×256) | $\phi_{tt} = 0.983\phi - 1.00\phi^3 + 0.100\phi_{xx} + 0.100\phi_{yy}$ | $\phi_{tt} = 1.010\phi - 0.998\phi^3 + 0.101\phi_{xx} + 0.101\phi_{yy}$ |
| 0% | $\phi_{tt} = 0.989\phi - 1.00\phi^3 + 0.100\phi_{xx} + 0.100\phi_{yy}$ | $\phi_{tt} = 1.575\phi - 0.827\phi^3 + 0.138\phi_{xx} + 0.135\phi_{yy}$ |
| 15% | $\phi_{tt} = 0.974\phi - 0.978\phi^3 + 0.100\phi_{xx} + 0.100\phi_{yy}$ | $\phi_{tt} = -90.8\phi - 264.4\phi^3 + 72.0\phi_{xx} + 67.7\phi_{yy} + 367.1\phi^2 + \mathbf{60.61}\phi_x \ldots \ldots$ |

**TABLE VIII.** Discovery of Single Klein-Gordon equation by ICNet and PiDL with different level noise.

| | Correct PDE $\phi_{tt} = \phi - \phi^3 + 0.1(\phi_{xx} + \phi_{yy})$ | |
|---|---|---|
| Noise | ICNet | PiDL |
| 25% | $\phi_{tt} = 0.980\phi - 0.979\phi^3 + 0.100\phi_{xx} + 0.100\phi_{yy}$ | $\phi_{tt} = 0.448\phi - 0.409\phi^3 + 0.100\phi_{xx} + 0.100\phi_{yy} + 0.409\phi^2$ |
| 50% | $\phi_{tt} = 0.901\phi - 0.937\phi^3 + 0.100\phi_{xx} + 0.099\phi_{yy}$ | $\phi_{tt} = -0.00\phi - 0.179\phi^3 + 0.100\phi_{xx} + 0.097\phi_{yy} + 0.780\phi^2 + \mathbf{0.007}\phi\phi_{yy} \ldots \ldots$ |

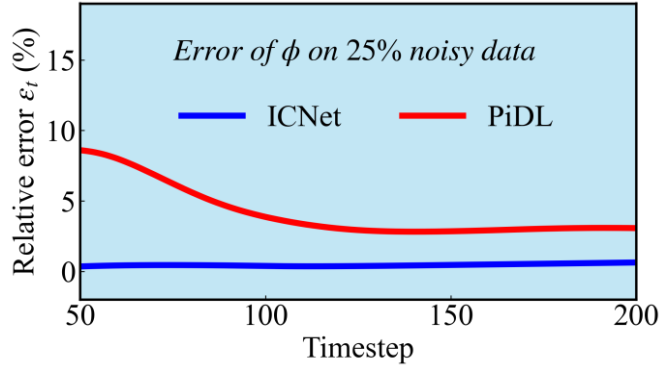

**FIG. 15.** Relative error $\varepsilon_t$ of ICNet and PiDL.

### 2. Case 5: Coupled Klein-Gordon equation

The coupled Klein-Gordon equation is used to further verify the superiority of the proposed method. The general form of coupled Klein-Gordon equation is as follows,

$$\phi_{1tt} = a_2 \phi_1 + b_2 (\phi_1^2 + \phi_2^2) \phi_1 + c_2 \Delta \phi_1$$

$$\phi_{2tt} = a_2 \phi_2 + b_2 (\phi_1^2 + \phi_2^2) \phi_2 + c_2 \Delta \phi_2 \quad (32)$$

where $\phi_1$ and $\phi_2$ denote the scalars of coupled scalar field, $a_2$, $b_2$, and $c_2$ are constant and $\Delta$ denotes the Laplace operator. Here, we set $a_2 = 1$, $b_2 = -1$ and $c_2 = 0.1$ to yield the datasets on the two-dimensional spatial domain $\Omega = [-\pi, \pi] \times [-\pi, \pi]$ with 256×256 identical mesh and time domain $t \in [0,4]$ with $\delta t = 0.01$. The following exponential initial condition is employed to simulate the coupled interaction of two scalar field

$$\phi_{10} = A_3 \times exp(-B_3((x - x_0)^2 + (y + y_0)^2)) - A_3 \times exp(-B_3((x + x_0)^2 + (y - y_0)^2))$$

$$\phi_{20} = A_3 \times exp(-B_3((x - x_0)^2 + (y - y_0)^2)) - A_3 \times exp(-B_3((x + x_0)^2 + (y + y_0)^2)) \quad (33)$$

Here we set $A_3 = 4$, $B_3 = 3$, $x_0 = 0.4$, and $y_0 = 0.4$.

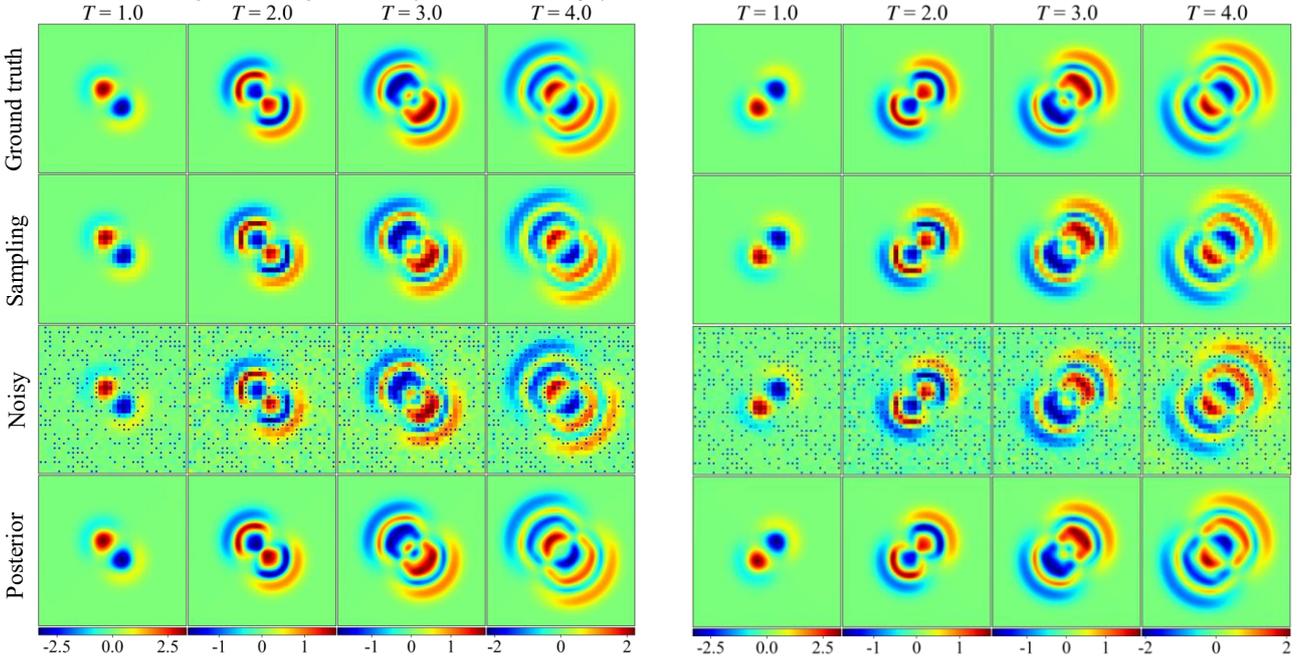

**FIG. 16.** Training data snapshots and posterior value snapshots of $\phi_1$ and $\phi_2$ for Coupled Klein-Gordon equation over two-dimensional domain $\Omega$, left is $\phi_1$ component and right is $\phi_2$ component.

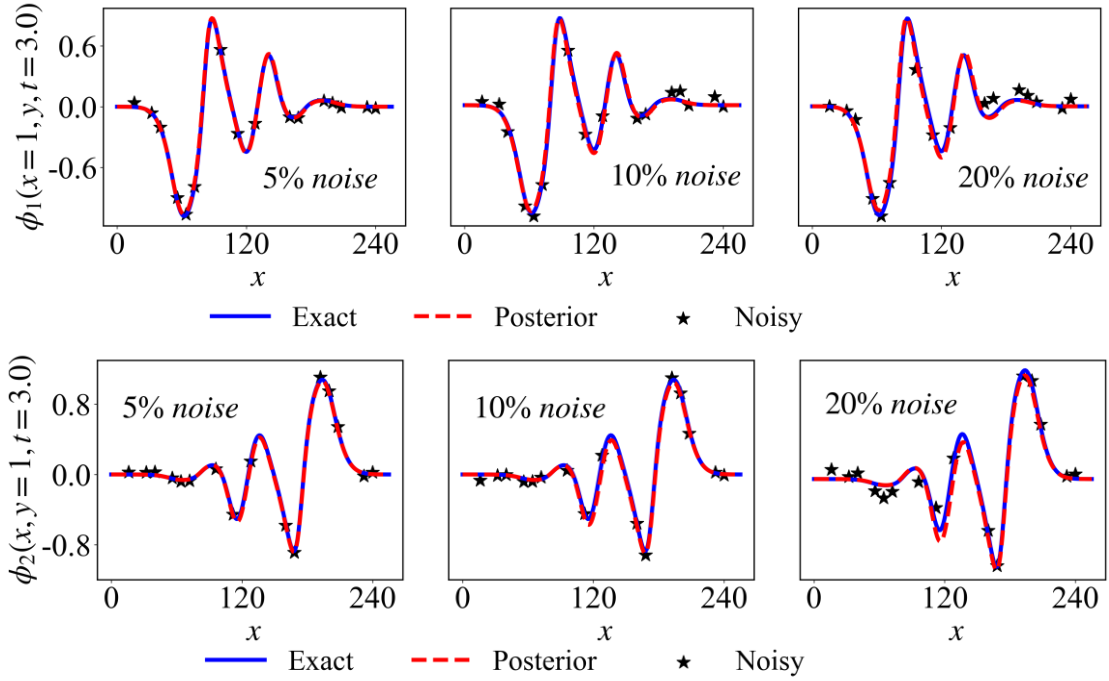

**FIG. 17.** Comparison among Exact data, Noisy data, and posterior data for Coupled Klein-Gordon equation at different level noise with limited data.

We also downsample the data from 256×256 mesh to 32×32 mesh with $\delta t = 0.01$, as shown in Fig. 16. A deep neural network with 10 hidden layers is constructed and per hidden layer has 60 neurons. The activation function of neural network employed here is tanh( • ). In the loss function, the weight coefficients $\alpha$ is set to be 1. $\beta$ is set to be $10^{-7}$ and $10^{-6}$ during the training process until the L-BFGS-B optimizer conergers. Meanwhile, the size of candidates remains unchanged. Adam optimizer is used to initialize the variables before the L-BFGS is adopted to accelerate the discovery. Fig. 17 shows the excellent matching degree of posterior data and true solution at different levels of noise with limited data. Fig. 18 shows the variation of coefficients and the loss function during the training process for Coupled Klein-Gordon equation with limited data at 20% noise level.

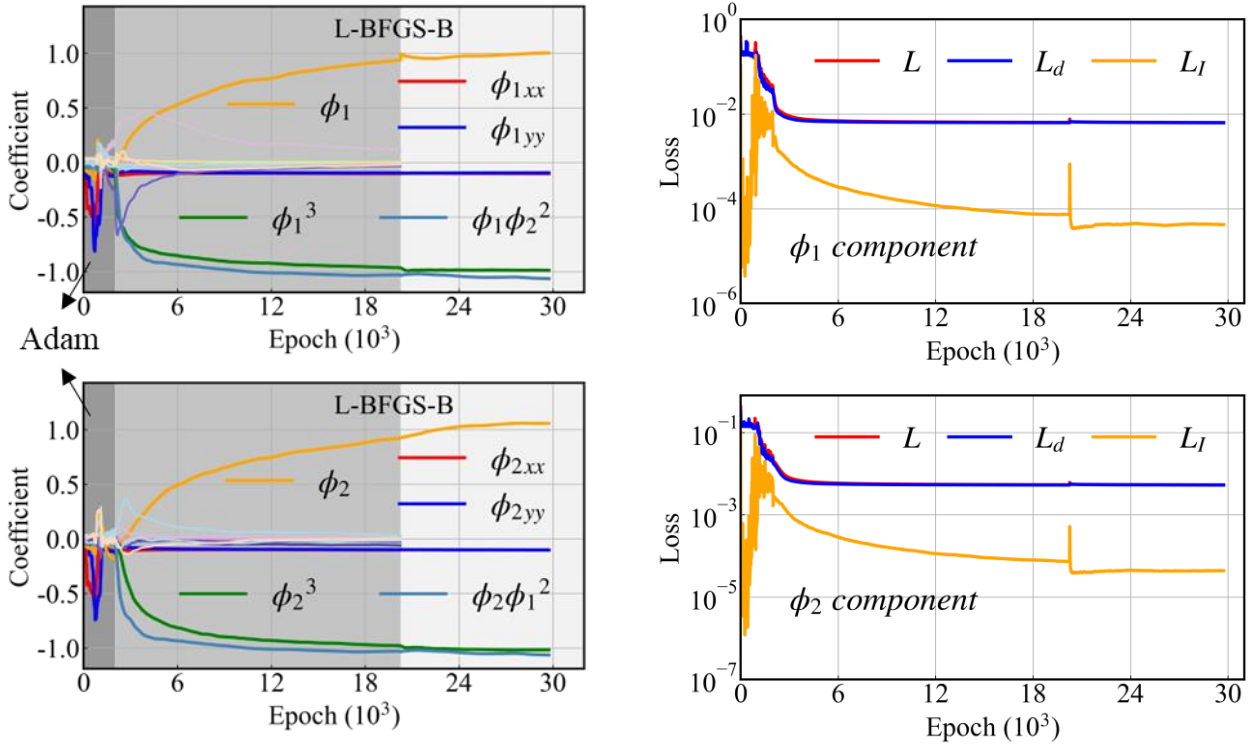

**FIG. 18.** Left is the coefficients variation along with the number of epochs for Coupled Klein-Gordon quation. The third part of the shaded area in the figure illustrates that $\|\Lambda\|_0$ remains unchanged. Except for the legends annotated in the figure, the remaining lines represent variations of other candidate terms. The right is the loss function of training process.

The comparison between ICNet and PDE-FIND is demonstrated first. PDE-FIND can still find the correct PDE with sparse data. However, if noise is added to the data, PDE-FIND fail to work, as shown in Table 9. To compare with PiDL, the performance of ICNet using less data is further investigated. Three hundred spatial points are randomly selected from the sparse data, which is about 1/3 of the sparse data (32×32) and 1/200 of the initial complete data (256×256). Simultaneously, 20 time-step data points are used for discovery. The comparison results are listed in Table 10. It can be seen that PiDL not only gives many redundant terms with large coefficients but also large error coefficients of active terms from limited data with noise. With an increase in the noise level, PiDL would fail to find correct active terms. However, ICNet can still discover the correct PDE. Fig. 19 demonstrates the advantages of ICNet compared with PiDL over the training data. Following the 110 candidates in the library used by PDE-FIND and PiDL,

$$\Theta = [1, \phi_{1x}, \phi_{1y}, \ldots, \phi_{2xy}, \phi_{2yy}, \phi_2, \phi_1, \ldots, \phi_1^2\phi_2, \phi_1^3, \phi_2\phi_{1x}, \phi_1\phi_{1x}, \cdots, \phi_1^2\phi_2\phi_{2yy}, \phi_1^3\phi_{2yy}] \quad (34)$$

However, there are following 13 candidates in the library embedded with Lorentz invariance

$$\Theta_L = [1, \phi_{1xx}, \phi_{1yy}, \phi_{2xx}, \phi_{2yy}, \phi_1, \phi_1^2, \phi_1^3, \phi_2^2\phi_1, \phi_2, \phi_2^2, \phi_2^3, \phi_1^2\phi_2] \quad (35)$$

**TABLE IX.** Discovery of Coupled Klein-Gordon equation by ICNet and PDE-FIND with different levels of noise.

| | Correct PDE | $\phi_{1tt} = \phi_1 - \phi_1^3 - \phi_1\phi_2^2 + 0.1(\phi_{1xx} + \phi_{1yy})$ | |
| | | $\phi_{2tt} = \phi_2 - \phi_2^3 - \phi_2\phi_1^2 + 0.1(\phi_{2xx} + \phi_{2yy})$ | |
|---|---|---|---|
| Noise | | ICNet | PDE-FIND |
| 0% (256×256) | | $\phi_{1tt} = 0.987\phi_1 - 0.998\phi_1^3 - 1.00\phi_1\phi_2^2$ $+0.100\phi_{1xx} + 0.100\phi_{1yy}$ $\phi_{2tt} = 0.978\phi_2 - 0.997\phi_2^3 - 0.997\phi_2\phi_1^2$ $+0.100\phi_{2xx} + 0.099\phi_{2yy}$ | $\phi_{1tt} = 1.00\phi_1 - 1.00\phi_1^3 - 1.00\phi_1\phi_2^2$ $+0.100\phi_{1xx} + 0.100\phi_{1yy}$ $\phi_{2tt} = 1.00\phi_2 - 1.00\phi_2^3 - 1.00\phi_2\phi_1^2$ $+0.100\phi_{2xx} + 0.100\phi_{2yy}$ |
| 0% | | $\phi_{1tt} = 0.968\phi_1 - 0.996\phi_1^3 - 0.997\phi_1\phi_2^2$ $+0.100\phi_{1xx} + 0.100\phi_{1yy}$ | $\phi_{1tt} = 1.275\phi_1 - 1.007\phi_1^3 - 1.001\phi_1\phi_2^2$ $+0.126\phi_{1xx} + 0.125\phi_{1yy}$ |

| 10% | $\phi_{2tt} = 0.962\phi_2 - 0.995\phi_2^3 - 0.994\phi_2\phi_1^2$ $+0.100\phi_{2xx} + 0.099\phi_{2yy}$ $\phi_{1tt} = 0.983\phi_1 - 0.998\phi_1^3 - 0.994\phi_1\phi_2^2$ $+0.100\phi_{1xx} + 0.098\phi_{1yy}$ $\phi_{2tt} = 0.955\phi_2 - 0.992\phi_2^3 - 0.992\phi_2\phi_1^2$ $+0.100\phi_{2xx} + 0.100\phi_{2yy}$ | $\phi_{2tt} = 1.271\phi_2 - 1.001\phi_2^3 - 0.992\phi_2\phi_1^2$ $+0.127\phi_{2xx} + 0.128\phi_{2yy}$ $\phi_{1tt} = 628.6\phi_1 - 67.4\phi_1^3 - 111.0\phi_1\phi_2^2$ $+100.4\phi_{1xx} + 98.87\phi_{1yy}$ $\mathbf{+480.3\phi_2\phi_{1y}} \dots \dots$ $\phi_{2tt} = 599.0\phi_2 - 89.6\phi_2^3 - 168.1\phi_2\phi_1^2$ $+100.9\phi_{2xx} + 92.7\phi_{2yy}$ $\mathbf{+538.3\phi_1\phi_{2y}} \dots \dots$ |
|---|---|---|

**TABLE X.** Discovery of Coupled Klein-Gordon equation by ICNet and PiDL with limited data at different levels of noise

| | Correct PDE | $\phi_{1tt} = \phi_1 - \phi_1^3 - \phi_1\phi_2^2 + 0.1(\phi_{1xx} + \phi_{1yy})$ $\phi_{2tt} = \phi_2 - \phi_2^3 - \phi_2\phi_1^2 + 0.1(\phi_{2xx} + \phi_{2yy})$ | |
|---|---|---|---|
| Noise | | ICNet | PiDL |
| 300 (0%) | | $\phi_{1tt} = 0.982\phi_1 - 0.987\phi_1^3 - 0.997\phi_1\phi_2^2$ $+0.100\phi_{1xx} + 0.098\phi_{1yy}$ $\phi_{2tt} = 0.961\phi_2 - 0.983\phi_2^3 - 0.987\phi_2\phi_1^2$ $+0.100\phi_{2xx} + 0.100\phi_{2yy}$ | $\phi_{1tt} = 0.811\phi_1 - 0.932\phi_1^3 - 0.741\phi_1\phi_2^2$ $+0.100\phi_{1xx} + 0.102\phi_{1yy}$ $\mathbf{+0.128\phi_1\phi_{2y}} \dots \dots$ $\phi_{2tt} = 0.769\phi_2 - 0.849\phi_2^3 - 0.831\phi_2\phi_1^2$ $+0.101\phi_{2xx} + 0.102\phi_{2yy}$ $\mathbf{+0.121\phi_2\phi_{1y}} \dots \dots$ |
| 300 (10%) | | $\phi_{1tt} = 1.014\phi_1 - 0.977\phi_1^3 - 1.030\phi_1\phi_2^2$ $+0.103\phi_{1xx} + 0.097\phi_{1yy}$ $\phi_{2tt} = 0.944\phi_2 - 0.974\phi_2^3 - 0.960\phi_2\phi_1^2$ $+0.100\phi_{2xx} + 0.100\phi_{2yy}$ | $\phi_{1tt} = 0.730\phi_1 - 0.900\phi_1^3 - 0.684\phi_1\phi_2^2$ $+0.096\phi_{1xx} + 0.090\phi_{1yy}$ $\mathbf{+0.138\phi_1\phi_{2y}} \dots \dots$ $\phi_{2tt} = 0.760\phi_2 - 0.836\phi_2^3 - 0.742\phi_2\phi_1^2$ $+0.096\phi_{2xx} + 0.099\phi_{2yy}$ $\mathbf{+0.099\phi_2\phi_{1x}} \dots \dots$ |
| 300 (20%) | | $\phi_{1tt} = 1.073\phi_1 - 0.986\phi_1^3 - 1.063\phi_1\phi_2^2$ $+0.102\phi_{1xx} + 0.094\phi_{1yy}$ $\phi_{2tt} = 1.059\phi_2 - 1.019\phi_2^3 - 0.944\phi_2\phi_1^2$ $+0.097\phi_{2xx} + 0.102\phi_{2yy}$ | $\phi_{1tt} = 0.00\phi_1 - 0.00\phi_1^3 - 0.00\phi_1\phi_2^2$ $+0.057\phi_{1xx} 0.035\phi_{1yy}$ $\mathbf{+0.089\phi_{1xy}} \dots \dots$ $\phi_{2tt} = 0.00\phi_2 - 0.00\phi_2^3 - 0.00\phi_2\phi_1^2$ $+0.058\phi_{2xx} + 0.035\phi_{2yy}$ $\mathbf{+0.090\phi_{2xy}} \dots \dots$ |

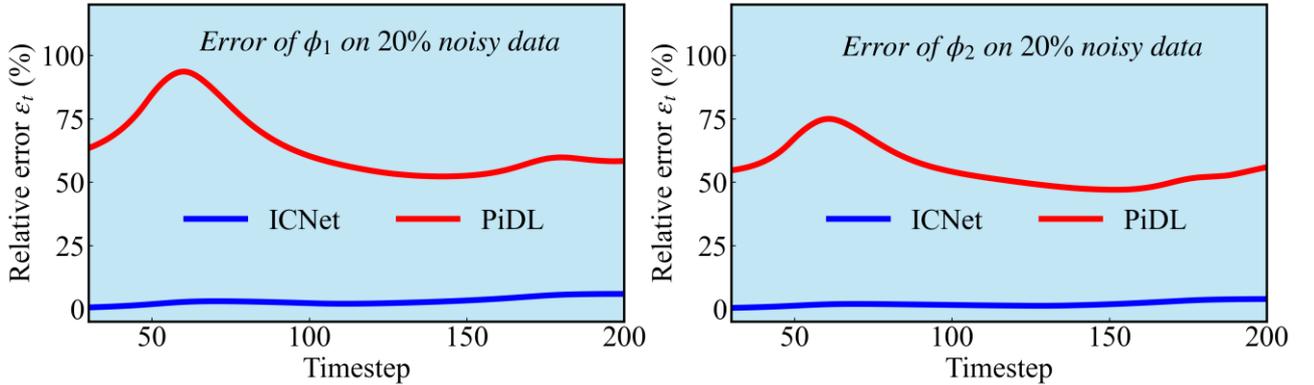

**FIG. 19.** Relative error $\varepsilon_t$ of ICNet and PiDL.

## IV. CONCLUSION

This paper proposes an ICNet method for embedding time and space translation invariance (Galilean and Lorentz invariance) into the discovery of governing equations. In this method, we firstly build ICNet by embedding the library satisfying the invariance into the loss function of neural network. We find the key point to perform the Galilean transformation and Lorentz transformation on the basic terms and their combinations. Specifically, in the PDE discovery of classical mechanics, the multidimensional vector **u** can only appear in $\mathbf{u} \cdot \nabla \mathbf{u}$, while the terms **u**, $\mathbf{u} \cdot \mathbf{u}$, $\mathbf{u} \cdot \mathbf{u} \cdot \nabla \mathbf{u}$, etc, cannot be included in the candidate terms. Similarly, in the context of relativity, the partial derivatives of the scalar u can only appear in the term $\nabla^2 u$, while other terms containing partial derivatives, such as $\nabla u$, $u \cdot \nabla^2 u$, etc., cannot be included in the candidate terms. Then we integrate the ICNet and STRidge to further decrease the equation residuals and improve the accuracy of discovered equation during the training process. At this stage, the invariance library is filtered gradually until the number of candidates no longer changes.

In this study, the efficacy of the ICNet method is validated using three governing equations of fluid mechanics (2D Burgers equation, the equation of Stenotic 2D channel flow over an obstacle, and the equation of 3D intracranial aneurysm). Then, ICNet is extended to the realm of relativity and validated through the 2D Single and Coupled Klein-Gordon equations. The ICNet method performs well with 50% noise and sparse data for the 2D Burgers equation, 2D Single and Coupled Klein-Gordon equation. The ICNet method also finds excellent results for equation of Stenotic 2D channel flow over an obstacle and equation of 3D intracranial aneurysm. Comparing the results obtained by PDE-FIND and PiDL, the ICNet method has better robustness in PDE discovery. Additionally, besides leveraging the automatic discretization technique of neural networks, the library with embedded invariance is another critical factor contributing to the superior performance of the ICNet method. Regardless of using the clean data without noise, PDE-FIND and PiDL still fail to discover realistic and reasonable results (N-S equation) for 2D channel flow over an obstacle and 3D intracranial aneurysms. Meanwhile, based on invariance library, integrating ICNet with STRidge without fixing the learnable coefficients during the training process enables ICNet to further decrease the equation residuals and discover high-quality equations.

## ACKNOWLEDGMENTS

We would like to acknowledge the Science Fund for Creative Research Groups of the National Natural Science Foundation of China (Grant No. 51921006), the National Natural Science Foundation of China (Grant Nos. 52108452), and Heilongjiang Touyan Team.

## AUTHOR DECLARATIONS

### Conflict of interest

We declare we have no competing interests.

### Author Contributions

**Chao Chen**: Investigation, Data curation, Methodology, Coding, Writing-original draft, Writing-review & editing. **Hui Li:** Main Idea, Investigation, Conceptualization, Methodology, Writing-original draft preparation, Writing-review & editing, Validation, Supervision, Funding acquisition. **Xiaowei Jin:** Investigation, Conceptualization, Validation, Supervision, Writing-original draft preparation, Writing-review & editing.

## DATA AVAILABLITY

All datasets and codes can be available by contacting the authors.

## APPENDIX A. COMPARISON OF EQUATION RESIDUALS WITH AND WITHOUT STRIDGE

### 1. Numerical examples for fluid mechanics

We first compare the equation residuals (eqrs) for governing equations of fluid mechnics over the training data and testing data, as shown in Fig.A.1, Fig.A.2, and Fig.A.3. It can be seen from the Fig.A.1 that the equation residuals for with (w) and without (w/o) STRidge are almost same. Because both of two equation have the identical PDE terms as same as true PDE. According to equation residuals over the testing data, we can judge the method whether discover the reasonable PDEs or the datasets exist other multiple solutions. It can be seen from the Fig.A.2 and Fig.A.3 that the equation residuals of ICNet without STRidge (ICNet w/o STRidge) are larger than ICNet with STRidge (ICNet w STRidge) and true PDE. This indicates that the existence of other redundant terms is not the multiple solution of the datasets. On the contrary, the existence of other terms would bring more error and are not reasonable. Also, the residuals of equation of ICNet w STRidge are closer to the true PDE, which further demonstrates the advantages of ICNet w STRidge based on invariance.

**TABLE XI.** Discovery of ICNet w STRidge and ICNet w/o STRidge for Burgers equation with 50% noise.

| Correct PDE | $u_t = -uu_x - vu_y + 0.1(u_{xx} + u_{yy})$ $v_t = -uv_x - vv_y + 0.1(v_{xx} + v_{yy})$ | |
|---|---|---|
| ICNet w STRidge | | ICNet w/o STRidge |
| $u_t = -1.001uu_x - 1.050vu_y + 0.103u_{xx} + 0.098u_{yy}$ $v_t = -0.899uv_x - 1.013vv_y + 0.098v_{xx} + 0.092v_{yy}$ | | $u_t = -0.996uu_x - 1.049vu_y + 0.103u_{xx} + 0.098u_{yy}$ $v_t = -0.870uv_x - 1.015vv_y + 0.097v_{xx} + 0.093v_{yy}$ |

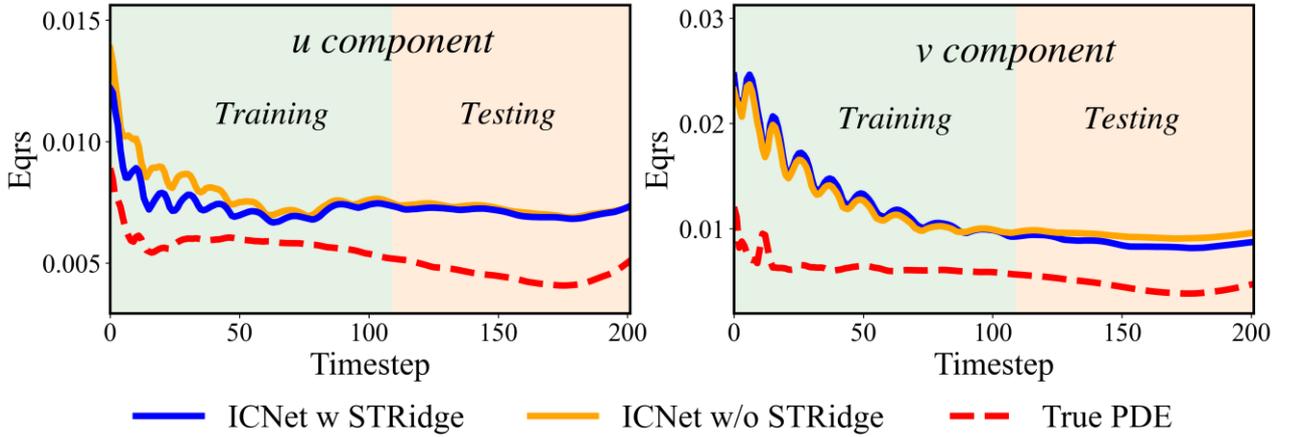

**FIG. 20.** Equation residuals of two-dimensional Burgers equation with and of ICNet w STRidge and ICNet w/o STRidge.

**TABLE XII.** Discovery of ICNet w STRidge and ICNet w/o STRidge for 2D Channel flow.

| Correct PDE | $u_t = -uu_x - vu_y - p_x + 0.2(u_{xx} + u_{yy})$ $v_t = -uv_x - vv_y - p_y + 0.2(v_{xx} + v_{yy})$ | |
|---|---|---|
| ICNet w STRidge | | ICNet w STRidge |
| $u_t = -0.999uu_x - 0.999vu_y - 1.00p_x$ $\quad +0.195u_{xx} + 0.195u_{yy}$ $v_t = -0.999uv_x - 0.999vv_y - 1.00p_y$ $\quad + 0.202v_{xx} + 0.201v_{yy}$ | | $u_t = -0.990uu_x - 990vu_y - 1.00p_x + 0.222u_{xx}$ $\quad +0.211u_{yy} + 1.274v_y + 1.240u_x \ldots$ $v_t = -0.990uv_x - 0.990vv_y - 1.00p_y + 0.213v_{xx}$ $\quad +0.169v_{yy} + 0.478v_{xy} + 0.478u_{xx} \ldots$ |

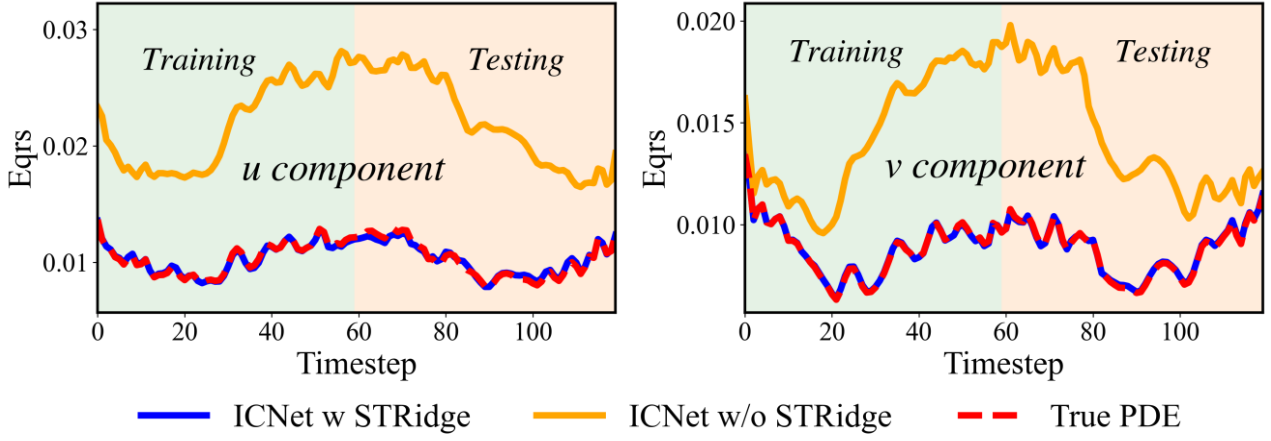

**FIG. 21.** Equation residuals of Stenotic 2D channel flow over an obstacle of ICNet w STRidge and ICNet w/o STRidge

**TABLE XIII.** Discovery of ICNet w STRidge and ICNet w/o STRidge for 3D intracranial aneurysm.

| Correct PDE | $u_t = -uu_x - vu_y - wu_z + p_x + 0.0102(u_{xx} + u_{yy} + u_{zz})$ $v_t = -uv_x - vv_y - wv_z + p_y + 0.0102(v_{xx} + v_{yy} + v_{zz})$ $w_t = -uw_x - vw_y - ww_z + p_z + 0.0102(w_{xx} + w_{yy} + w_{zz})$ | |
|---|---|---|
| ICNet w STRidge | | ICNet w/o STRidge |
| $u_t = -0.955uu_x - 0.955vu_y - 0.955wu_z + 1.00p_x$ $\quad + 0.0100u_{xx} + 0.0110u_{yy} + 0.0097u_{zz}$ $v_t = -0.955uv_x - 0.955vv_y - 0.955wv_z + 1.00p_y$ $\quad + 0.0097v_{xx} + 0.0107v_{yy} + 0.0097v_{zz}$ $w_t = -0.955uw_x - 0.955vw_y - 0.955ww_z + 1.00p_z$ $\quad + 0.0100w_{xx} + 0.0099w_{yy} + 0.0114w_{zz}$ | | $u_t = -0.891uu_x - 891vu_y - 891wu_z + 1.00p_x$ $\quad + 0.009u_{xx} + 0.009u_{yy} + 0.010u_{zz}$ $\quad -0.156u_x - 0.152v_y \ldots \ldots$ $v_t = -891uv_x - 891vv_y - 891wv_z + 1.00p_y$ $\quad + 0.008v_{xx} + 0.005v_{yy} + 0.010v_{zz}$ $\quad +0.147w_z + 0.145u_x \ldots \ldots$ $w_t = -891uw_x - 891vw_y - 891ww_z + 1.00p_z$ $\quad + 0.009w_{xx} + 0.009w_{yy} + 0.006w_{zz}$ $\quad +0.071v_y + 0.065w_z \ldots \ldots$ |

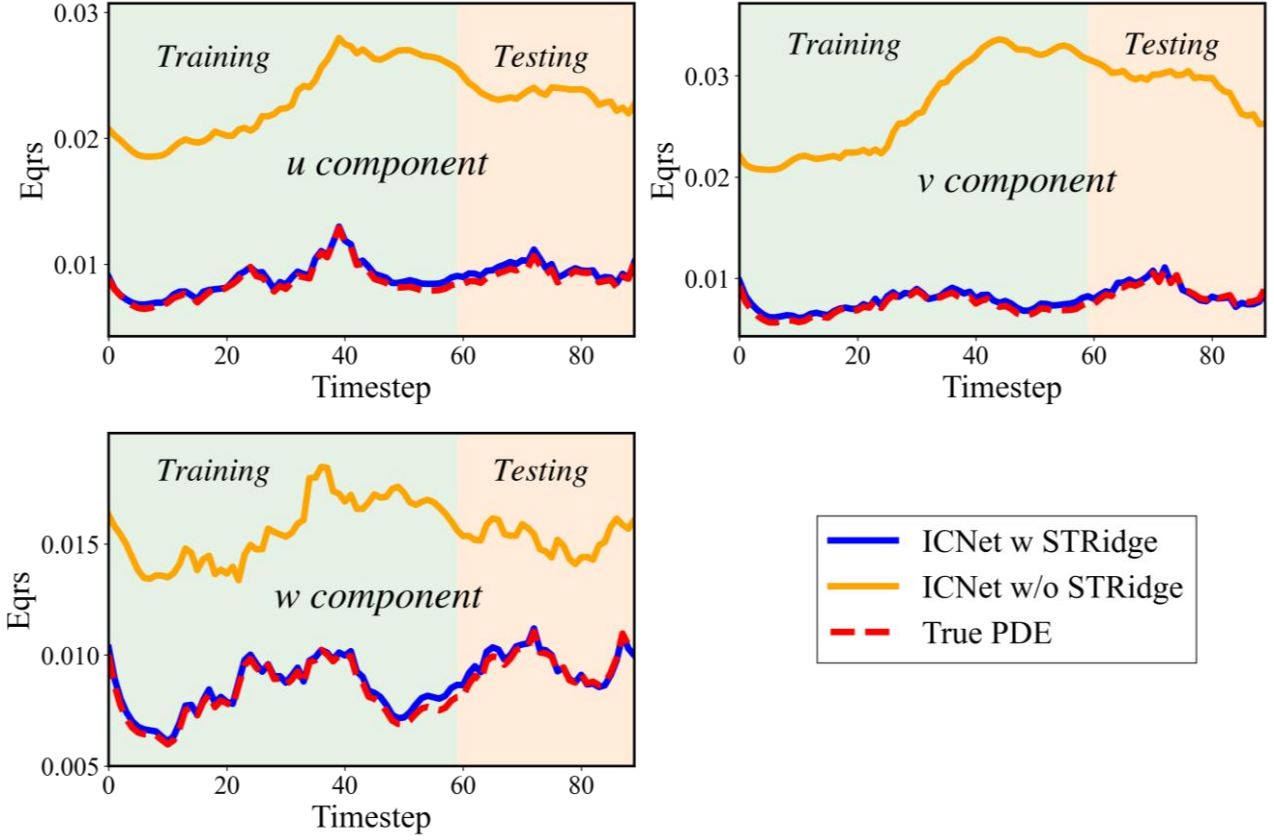

**FIG. 22.** Equation residuals of three-dimensional intracranial aneurysm of ICNet w STRidge and ICNet w/o STRidge.

## 2. Numerical examples for relativity

We also compare the equation residuals (eqrs) for wave equations of relativity over the training data and testing data, as shown in Fig.A.4 and Fig.A.5. It can be seen from the figure that the equation residuals for with and without STRidge are almost same. Because both of two equation have the identical PDE terms as same as true PDE.

**TABLE XIV.** Discovery of ICNet w STRidge and ICNet w/o STRidge for Single Klein-Gordon equation with 50% noise.

| Correct PDE | $\phi_{tt} = \phi - \phi^3 + 0.1(\phi_{xx} + \phi_{yy})$ |
|---|---|
| ICNet w STRidge | ICNet w/o STRidge |
| $\phi_{tt} = 0.901\phi - 0.937\phi^3 + 0.100\phi_{xx} + 0.099\phi_{yy}$ | $\phi_{tt} = 0.896\phi - 0.930\phi^3 + 0.100\phi_{xx} + 0.099\phi_{yy}$ |

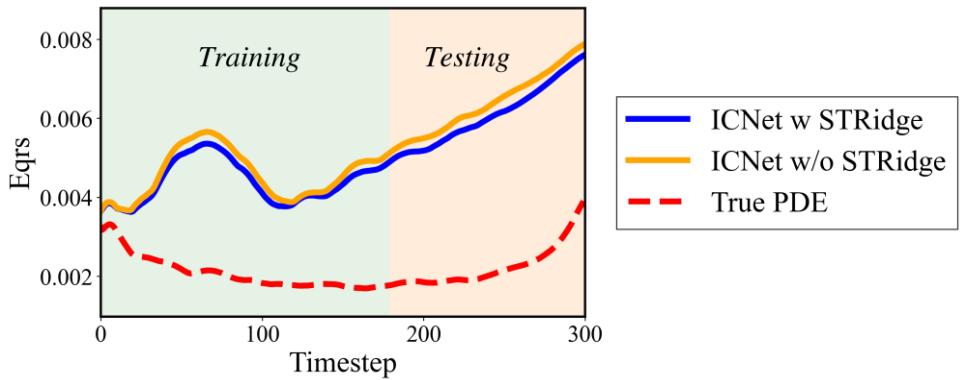

**FIG. 23.** Equation residuals of Single-Klein-Gordon equation of ICNet w STRidge and ICNet w/o STRidge

**TABLE XV.** Discovery of ICNet with STRidge and ICNet without STRidge for Coupled Klein-Gordon equation with limited data at 20% noise

| Correct PDE | $\phi_{1tt} = \phi_1 - \phi_1^3 - \phi_1\phi_2^2 + 0.1(\phi_{1xx} + \phi_{1yy})$ |
| --- | --- |
| | $\phi_{2tt} = \phi_2 - \phi_2^3 - \phi_2\phi_1^2 + 0.1(\phi_{2xx} + \phi_{2yy})$ |

| ICNet w STRidge | ICNet w/o STRidge |
| --- | --- |
| $\phi_{1tt} = 1.073\phi_1 - 0.986\phi_1^3 - 1.063\phi_1\phi_2^2$ $+0.102\phi_{1xx} + 0.094\phi_{1yy}$ | $\phi_{1tt} = 0.994\phi_1 - 0.969\phi_1^3 - 1.036\phi_1\phi_2^2$ $+0.102\phi_{1xx} + 0.095\phi_{1yy}$ |
| $\phi_{2tt} = 1.059\phi_2 - 1.019\phi_2^3 - 0.944\phi_2\phi_1^2$ $+0.097\phi_{2xx} + 0.102\phi_{2yy}$ | $\phi_{2tt} = 0.927\phi_2 - 0.982\phi_2^3 - 0.926\phi_2\phi_1^2$ $+0.098\phi_{2xx} + 0.099\phi_{2yy}$ |

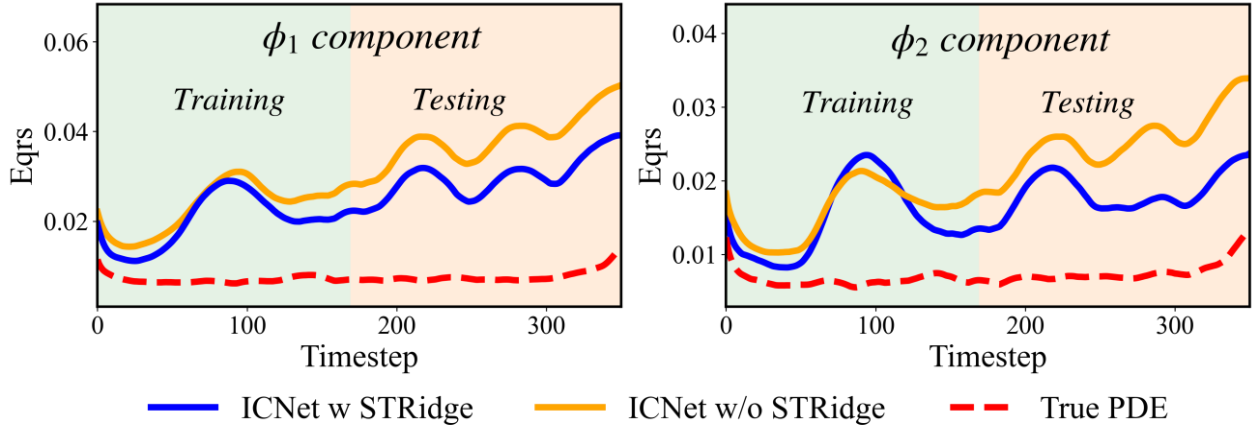

**FIG. 24.** Equation residuals of coupled Klein-Gordon equation of ICNet w STRidge and ICNet w/o STRidge